\pdfoutput=1

\documentclass[11pt]{article}

\usepackage{acl}

\usepackage{times}
\usepackage{latexsym}

\usepackage[T1]{fontenc}

\usepackage[utf8]{inputenc}

\usepackage{microtype}

\usepackage{comment}
\usepackage{bm}
\usepackage{graphicx}
\usepackage{multirow}
\usepackage{booktabs}
\usepackage{algorithm}
\usepackage{algpseudocode}
\usepackage{amsmath}
\usepackage{tikz}
\usepackage{pgfplots}
\usepackage{xcolor}
\usepackage{colortbl}
\usepackage{xspace}
\usepackage{tabularx}
\usepackage{wrapfig}

\title{UNcommonsense Reasoning:\\Abductive Reasoning about Uncommon Situations}

\author{Wenting Zhao$^{1}$\Thanks{\ Wenting, Lorraine, and Alane's work done at AI2.
Lorraine and Alane are co-last authors.} 
\quad Justin T. Chiu$^{1}$ 
\quad Jena Hwang$^{2}$ 
\quad Faeze Brahman$^{2}$  
\quad Jack Hessel$^{2}$ \\
\bf \quad Sanjiban Choudhury$^{1}$ 
\quad Yejin Choi$^{2, 3}$
\quad Xiang Lorraine Li$^{4 *}$ \quad Alane Suhr$^{5 *}$\\\\
$^1$Cornell University, $^2$Allen Institute for Artificial Intelligence \\
$^3$University of Washington, $^4$University of Pittsburgh, $^5$University of California, Berkeley \\ 
\texttt{wz346@cornell.edu, xianglli@pitt.edu, suhr@berkeley.edu}}

\newcommand{\as}[1]{{\color{teal} [Alane: {#1}]}}

\newcommand{\jack}[1]{{\color{blue} [Jack: {#1}]}}
\newcommand\fb[1]{{\color{brown}[\textit{#1}]$_{-FB}$}}

\newcommand{\ignore}[1]{}

\newcommand{\stdev}[2]{#1 \begin{small}$\pm$ #2\end{small}}
\newcommand{\corpus}{\textsc{UNcommonsense}\xspace}

\newcommand{\story}{{\textbf{un}-RocStories}\xspace}
\newcommand{\social}{\textbf{un}-{SocialIQA}\xspace}

\newcommand{\llm}{{\emph{LLM}}\xspace}

\newcommand{\llmhuman}{{\emph{C+LLM}}\xspace}
\newcommand{\human}{\emph{Crowd}\xspace}

\usepackage{titlesec}
\titlespacing*{\section}{0pt}{1ex}{1ex}
\titlespacing*{\subsection}{0pt}{1ex}{1ex}

\begin{document}
\maketitle

\begin{abstract}
Language technologies that accurately model the dynamics of events must perform commonsense reasoning.
Existing work evaluating commonsense reasoning focuses on making inferences about common, everyday situations.
To instead investigate the ability to model \textbf{un}usual, \textbf{un}expected, and \textbf{un}likely situations, we explore the task of \textbf{un}commonsense abductive reasoning. 
Given a piece of context with an unexpected outcome, this task requires reasoning abductively to generate an explanation that makes the unexpected outcome more likely in the context.
To this end, we curate and release a new English language corpus called \textbf{UNcommonsense}. 
We characterize the performance differences between human explainers and the best-performing large language models, finding that model-enhanced human-written explanations achieve the highest quality by trading off between specificity and diversity. 
Finally, we experiment with several imitation learning algorithms to train open and accessible language models on this task. 
When compared with the vanilla supervised fine-tuning approach, these methods consistently reduce lose rates on both common and uncommonsense abductive reasoning judged by human evaluators.

\end{abstract}

%

\section{Introduction}

The ability to perform commonsense reasoning is crucial for understanding the dynamics of everyday events, both for humans and for natural language processing systems. 
However, most existing commonsense reasoning benchmarks focus on the ability to model common events~\cite{sap-etal-2019-social,talmor-etal-2019-commonsenseqa,lin-etal-2020-commongen}, i.e., {\em given a commonly encountered situation, what commonsense inferences can be made?}
Comparatively less effort has been devoted to evaluating a different class of inputs: unusual scenarios, improbable situations, and implausible events.

Understanding and reasoning about these situations is crucial for the fairness and reliability of language technologies. For example, most LLMs are trained on English data. They are accustomed to Western cultural norms, and therefore non-English culture could be considered uncommon in current LLM-based NLP systems, e.g., wearing shoes indoors is normal in Western culture, but is often viewed as disrespectful in Asian households. Being able to reason about uncommon situations helps LLMs serve individuals from diverse cultural backgrounds more effectively. Uncommon situations could also be associated with important and high-risk scenarios~\cite{weidinger2022taxonomy}. Consider a situation where an individual tries out a massage chair and subsequently develops small, itchy, and red welts on their back. One explanation may be that this person is allergic to vibrations, a rare yet real medical condition called vibratory urticaria. While this is an uncommon situation, an NLP system that incorrectly interprets or handles this situation could lead to severe consequences, for example a misdiagnosis of a more common condition unrelated to the chair.

\begin{figure}[t!]
    \centering
    \includegraphics[width=8cm]{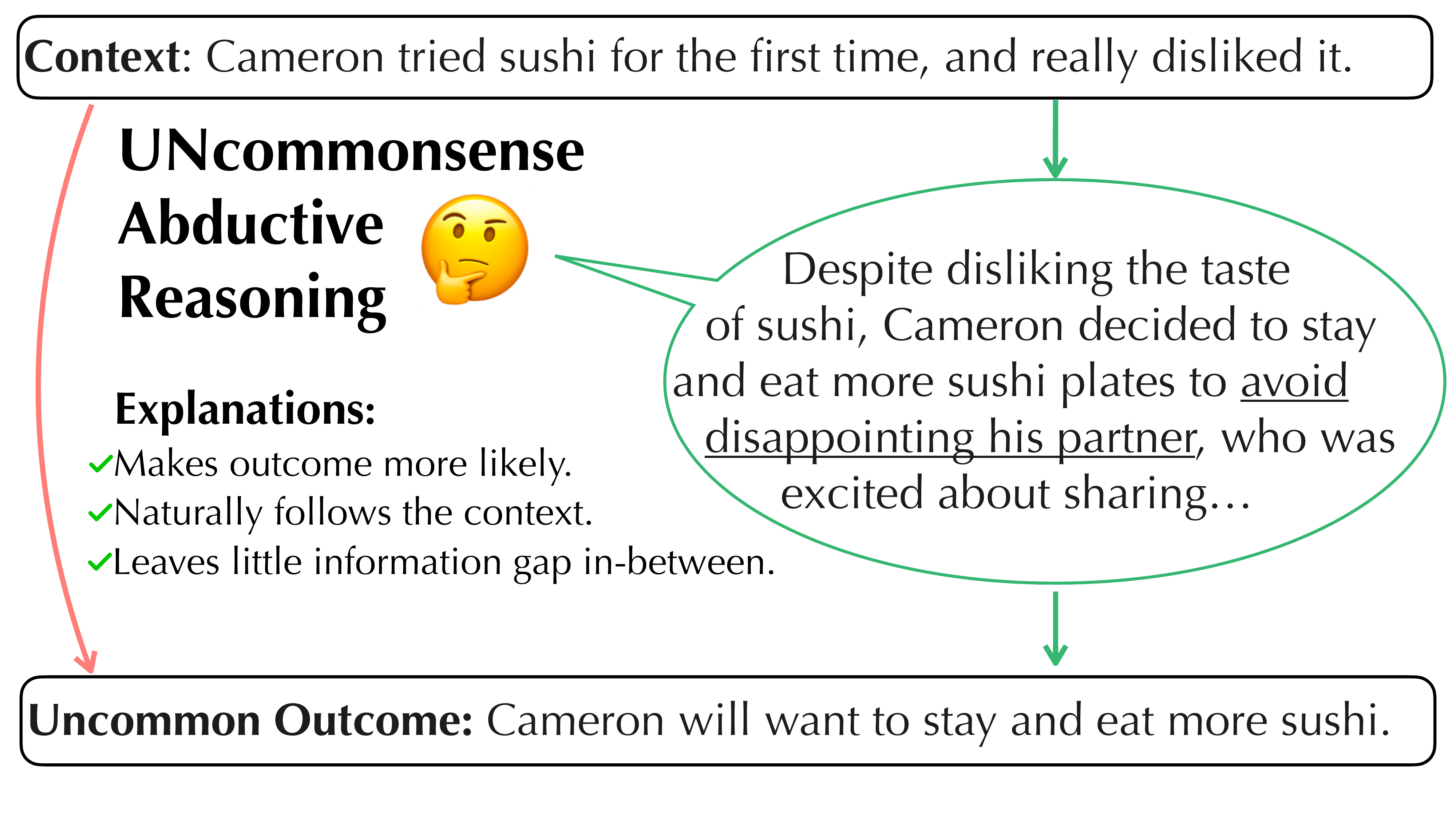}
    \vspace{-20pt}
    \caption{Given a context and an uncommon outcome, uncommonsense abductive reasoning aims to produce an explanation so that the unlikely outcome becomes likely. The explanation needs to follow the three rules noted with the check marks. 
    }
    \label{fig:intro}
\end{figure}

To bridge this gap, we introduce \corpus, a benchmark that explicitly challenges models to reason about implausible, yet still possible, events. 
\corpus is an English-language corpus consisting of 20k unique contexts paired with explicitly uncommon outcomes. 
We source uncommon outcomes from the incorrect answers in several multiple choice commonsense reasoning  benchmarks, which were designed to challenge models to identify the most likely outcome among multiple candidates, given a context. 
Given these contexts and uncommon outcomes, we crowdsource 41k abductive explanations, which provide a plausible explanation of how an uncommon outcome could have arisen, given an input context. 
See Figure \ref{fig:intro} for an example. \textsc{UNcommonsense} complements existing commonsense reasoning corpora
~\cite[e.g., ][]{mostafazadeh-etal-2016-corpus,Bhagavatula2020Abductive,rudinger-etal-2020-thinking} that focus on reasoning about common events.\footnote{Data is available at \url{huggingface.co/datasets/allenai/UNcommonsense}}

We examine the gap between human and model performance in generating abductive uncommonsense explanations, finding subtle differences in explanation quality. 
Given a few demonstrations, the top-performing LLM GPT-4~\cite{openai2023gpt4} produces more specific explanations than those acquired through crowdsourcing; however, these explanations are less diverse. 
While their explanations often lack sufficient details to connect contexts to outcomes, workers recruited through crowdsourcing excel at creating a broader picture of possible intermediate events.
To combine the creativity of human authors and the specificity of LLM-generated explanations, we experiment with using an LLM to refine crowd-authored explanations by filling in more details. 
Though LLM-generated explanations are generally preferred over the original crowd-written explanations, we find that LLM-refined crowd-written explanations hold a notable advantage over those generated only by an LLM.


Generating abductive explanations for uncommon outcomes \emph{without} conditioning on a human-written starting point remains a challenge, particularly for publicly available models.
Specifically, we find that the purely offline learning approach of supervised fine-tuned models suffer from compounding errors during generation.
This is particularly problematic for our task, which generally requires lengthy explanations that bridge the gap between a context and an uncommon outcome.
To this end, we experiment with two online imitation learning methods to improve the performance of open and accessible language models on abductive reasoning.
When compared with supervised fine-tuning, these methods show an absolute 10\% increase in win rates against the strong GPT-4 baseline when evaluated by workers on both commonsense and uncommonsense abductive reasoning.


\begin{table*}[]
\footnotesize
\centering
\begin{tabular}{p{0.35\textwidth} p{0.25\textwidth} p{0.3\textwidth}}
\toprule
  \multicolumn{1}{c}{Context} &
  \multicolumn{1}{c}{Uncommon Outcome} &
  \multicolumn{1}{c}{Explanation} \\ \midrule
  Kai bought a Kindle from Amazon and used it all of the time. &
  Kai will want to return the Kindle and go back to reading physical books only. &
  After a month of reading books with the Kindle, the free book trial ran out and Kai decided that reading books on the Kindle was not worth paying for. The return period for the Kindle has not ended yet. \\ \midrule
  Tracy went shopping at the market and brought many good items at the super market like fish and meat. &
  Tracy will want to get angry. &
  Tracy realized that many of the items she bought were already expired, and the shopkeepers had knowingly sold her expired meats. \\ \midrule
  Scott was hungry. He decided to cook dinner. He cooked tacos. He made enough to share with a friend. &
  His friend was so offended he asked Scott to leave. &
  Scott made the tacos with beef and didn't tell his friend until after they ate, even though he know that his friend was a strict vegetarian. \\ \midrule
  Drew order a small ice cream cone at the Drive Thru. He drove to the front window to wait for his order. The cashier handed the ice cream with one of her hands. Drew took the ice cream cone and turned it upside down. &
  Drew kept his car clean this way. &
  He just dumped the whole thing into a small plastic cup he kept in the car and then he ate it out of the cup. \\
\bottomrule
\end{tabular}
\caption{\corpus examples. The first two examples are from \social and the next two examples come from \story; explanations are written by crowdworkers.
}
\label{tab:human-example}
\end{table*}

\section{Uncommonsense Abductive Reasoning}

Given a natural language context $x$ and outcome $y$, the task of abductive reasoning requires generating a natural language explanation $z$ that augments the context, making the outcome more probable~\cite{Bhagavatula2020Abductive}.
In uncommonsense abductive reasoning, we focus on situations where an outcome $y$ is very unlikely to happen in context $x$.
For example, in Figure~\ref{fig:intro}, our context ``\emph{Cameron tried sushi for the first time, and really disliked it.}'' is paired with the unlikely outcome ``\emph{Cameron will want to stay and eat more sushi.}''.
One possible abductive explanation of this outcome is that ``\emph{... Cameron decided to stay and eat more sushi plates to avoid disappointing his partner, who was excited about sharing...}''.
When the context is augmented with this explanation, it becomes significantly more likely that the outcome will occur.

To our knowledge, no existing datasets explicitly study abductive reasoning for uncommon situations. 
We fill this gap by collecting the \corpus dataset, which contains contexts paired with both uncommon outcomes and explanations that rationalize these uncommon outcomes.
Table~\ref{tab:human-example} presents several examples from \corpus, with explanations written by humans. 
In this section, we describe our process for collecting \corpus, including collecting uncommon outcomes and abductive explanations.

\subsection{Uncommon Outcomes}
\label{sec:outcomes}
We first collect pairs of contexts and uncommon outcomes.
We source contexts from two existing commonsense datasets: SocialIQA~\cite{sap-etal-2019-social} and ROCStories~\cite{mostafazadeh-etal-2016-caters}.
Each uncommon outcome is either human-written or LLM-generated. 

\paragraph{\social.} SocialIQA is a multiple-choice question answering dataset created to evaluate reasoning about social interactions. 
Each example consists of a context $x$, a question $q$, and three answer choices $\mathcal{A}$, one of which is correct. To pick the uncommon outcome, we identify the least likely answer choice (among the incorrect ones) by we computing $\texttt{argmin}_{a\in \mathcal{A}^-} p(a|x, q)$ with GPT-3, where $\mathcal{A}^-$ is the set of two incorrect answers. We then use LLM prompting\footnote{All prompting templates can be found in Appendix~\ref{sec:template}.} to combine the question and the least likely \emph{incorrect} answer choice into a declarative sentence, which we take as the uncommon outcome $y$.


All original SocialIQA answer choices are human-written.
To further diversify uncommon outcomes, we additionally generate new improbable answer choices using few-shot prompting with LLMs.
We use 6-shot prompting with GPT-4\footnote{We use gpt4-0314 for all generation tasks, including uncommon outcomes, explanations, and during online learning.} to produce one improbable answer for a randomly sampled subset of SocialIQA contexts and questions, then combine the question and generated answer into into uncommon outcomes using the same procedure above. 

\paragraph{\story.} The ROCStories Cloze Test includes examples of four-sentence stories paired with two sentence-length endings.
The original task is to predict which of the two endings is more likely.
In \corpus, we take each four-sentence story as the context $x$ and the \emph{incorrect} ending as the uncommon outcome $y$.

\paragraph{Filtering out common outcomes.}
To focus on uncommon scenarios, we exclude examples where outcomes are obvious in the context.\footnote{Both human-written and LLM-generated outcomes can be too obvious without filtering.} 
We prompt GPT-4 to rate the likelihood of the outcome given the context on a scale from 1 to 5, and remove examples with ratings of 4 or 5. 
Filtering with this criterion removes 0.7\% of \story examples and 1.82\% of \social examples.
\subsection{Explanations for Uncommon Outcomes}
\label{sec:explanations}
We crowdsource explanations of uncommon outcomes $z$ on Amazon Mechanical Turk (MTurk) from 156 unique workers, with a pay rate of 15 USD/hour.\footnote{Appendix~\ref{app:crowdsourcing} contains additional details on crowdsourcing.}
We also experiment with using an LLM both to generate explanations from scratch given contexts paired with uncommon outcomes, and to enhance crowd-written explanations. Specifically, we use GPT-4, which has demonstrated strong reasoning abilities on a wide range of tasks.

\paragraph{Explanation Writing.}
We first conduct a paid qualification task that identifies 204 workers who write high-quality explanations, who are then invited to participate in explanation writing tasks.
Tasks are launched in small batches, and we evenly distribute tasks across workers in each batch, which, by design, ensures that no worker writes too many explanations.
Due to the subjectivity on evaluation for this task, we emphasize collecting a wide variety of explanations on the development and test sets, creating no less than three tasks for each pair of context and outcome collected in Section~\ref{sec:outcomes}.
We also perform extensive quality control on collected explanations, described in Appendix~\ref{app:crowdsourcing}.
We also use this task to identify the outcomes that are impossible given their contexts, asking workers to mark these examples and provide their reasoning.
We remove examples marked as impossible by more than half of its annotators.

\paragraph{LLM-Enhanced Crowd-written Explanations.}
We prompt LLMs to enhance crowd-written explanations.
We instruct GPT-4 to add details that better connect contexts and outcomes.
\paragraph{LLM-Generated Explanations.}
We use 3-shot prompting with GPT-4 to generate explanations for each context-outcome pair. 

\paragraph{LLM-Enhanced LLM-Generated Explanations.}
To directly investigate the effect of LLM-based explanation enhancement, we also apply LLM enhancement to one randomly-chosen \emph{LLM} explanation for each context-outcome pair, using the same prompting method that was used to enhance \human explanations.
We refer to these LLM-enhanced LLM-generated explanations as \emph{LLM$^2$}.

\section{Data Analysis}
\label{sec:analysis}

\begin{table}[]
    \centering \footnotesize
    \begin{tabular}{lcc} \toprule
         & \story & \social \\ \midrule
       \multicolumn{3}{l}{\textit{\# of context-outcome $(x, y)$ pairs, with $y$ sourced from...}} \\
       Human & 1,775 / \phantom{0,}765 / \phantom{0,}999 & \phantom{0}5,531 / \phantom{0,}543 / \phantom{0,}999 \\
       LLM & \phantom{0,00}0 / \phantom{0,00}0 / \phantom{0,00}0  & \phantom{0}8,699 / \phantom{0,}931 / \phantom{0,}705 \\ \midrule
       \multicolumn{3}{l}{\textit{\# of explanations $z$, sourced from...}} \\
        \emph{Crowd} & \phantom{0}8,428 / 4,240 / 4,835 & 14,563 / 4,407/ 5,238 \\
        \emph{C+LLM} &  \phantom{0}8,333 / 4,203 / 4,771 & 14,469 / 4,390 / 5,209\\ 
        \emph{LLM} & 17,548 / 7,556 / 9,919 & 14,324 / 4,422 / 5,112\\ \bottomrule
    \end{tabular}
    \caption{Basic statistics of \corpus. Counts in cells report the number of examples split across the train/dev/test sets.}
    \label{tab:dataset-stats}
    \vspace{-10pt}
\end{table}
Table~\ref{tab:dataset-stats} contains basic statistics of the collected data. 
\corpus includes 3,539 contexts paired with uncommon outcomes in \story and 17,408 in \social for a total of 20,947 context-outcome pairs.
We adopt the same train/dev/test splits as the original releases of RocStories and SocialIQA.
In total, we collect 41,711 crowd-written explanations (\emph{Crowd}), 41,375 LLM-enhanced crowd-written explanations (\emph{C+LLM}), and 58,881 LLM-generated explanations (\emph{LLM}).
We compare explanations from these three sources using several metrics, including human preference judgments, explanation lengths, and measures of explanation diversity.

\begin{table}[t]
\centering
\footnotesize
\begin{tabular}{ccccc}
\toprule
& \multicolumn{3}{c}{\corpus} & $\alpha$NLG\\
  $l$ & \story & \multicolumn{2}{c}{\social} & \\
  & & (Human) & (LLM) & \\ \midrule
5 & \phantom{0}0.0   & \phantom{0}0.0   & \phantom{0}0.0  & \phantom{0}0.1  \\
4 & \phantom{0}0.0   & \phantom{0}0.0  & \phantom{0}0.0  & 31.8 \\
3 & 29.4 & 50.7 & 25.8 & 40.3 \\
2 & 63.1 & 42.1 & 59.6 & 19.9 \\
1 & \phantom{0}7.5  & \phantom{0}6.9 & 14.5 & \phantom{0}0.9 \\
\bottomrule
\end{tabular}
\caption{Proportion of outcomes assigned likelihoods $l \in \{1\dots5\}$ for examples in \corpus corresponding to \story and \social (split by human-authored and LLM-generated uncommon outcomes), compared with $\alpha$NLG.
}
\label{tab:outcome-likelihood-by-dataset}
\end{table}

\paragraph{Unlikely Outcomes.}
We utilize GPT-4 prompting to quantify, on a scale from 1 to 5, how likely an outcome may occur given the context.
Table~\ref{tab:outcome-likelihood-by-dataset} summarizes the ratios of outcomes broken down by their scales with 1 being the most unlikely.
In $\alpha$NLG, only 20.8\% of outcomes have a scale of 1 or 2.
Significantly more outcomes are rated 1 or 2 in \story (70.6\% of outcomes) and \social (49.0\% of human-written and 74.1\% of LLM-generated outcomes).
Compared to $\alpha$NLG, \corpus poses a unique challenge of abductive reasoning about uncommon outcomes.

\paragraph{Explanation Preferences.}

\begin{table*}[htbp]
    \centering
    \begin{minipage}[b]{.66\textwidth}
        \centering
        \footnotesize
        \begin{tabular}{lcccccc}
\toprule
     & \multicolumn{3}{c}{\social} & \multicolumn{3}{c}{\story} \\ 
     & \emph{Crowd}     & \emph{C+LLM}     & \emph{LLM$^2$}    & \emph{Crowd}     & \emph{C+LLM}     & \emph{LLM$^2$}      \\ \midrule
Win  & 30.8	&43.2	&33.8	&19.2	&28.4	&26.4           \\
Eql. good  & 33.4	&34.8	&41.2	&37.0	&45.6	&42.4         \\
Eql. bad  & \phantom{0}3.4	&\phantom{0}2.0	&\phantom{0}3.8	&12.0	&\phantom{0}3.0	&\phantom{0}3.0      \\ 
Lose & 32.4	&20.0	&21.2	&42.6	&23.0	&28.2         \\\midrule
Fleiss' $\kappa$ &0.47&0.47&0.55&0.48&0.43&0.53\\
\bottomrule
\end{tabular}
        \caption{Preference judgments given by crowdworkers comparing explanations from \emph{LLM} with explanations from \emph{Crowd}, \emph{C+LLM}, and \emph{LLM$^2$}.}
        \label{tab:win-rate-human-gpt4}
    \end{minipage}%
    \hfill
    \begin{minipage}[b]{.32\textwidth}
        \centering
        \footnotesize
        \begin{tabular}{lcc}
\toprule
$l$ & \social &\story \\ \midrule
2 & 77.35 & 71.63 \\
1 & 90.90 & 75.85 \\
\bottomrule
\end{tabular}
        \caption{Non-lose rates of \llmhuman versus \llm, broken down by the likelihoods $l$ of outcomes ($l=2$ is more likely, and $l=1$ is less likely). 
}
        \label{tab:breakdown-by-likelihood}
    \end{minipage}
    \vspace{-10pt}
\end{table*}



We first compare pairwise preferences of \emph{LLM} explanations versus \emph{Crowd}, \emph{C+LLM}, and \emph{LLM$^2$} explanations.
We randomly sample 500 context-outcome pairs from each \corpus test set, and select the same explanation from \emph{LLM} that was randomly chosen to be enhanced into \emph{LLM$^2$}.
We then randomly sample a single crowd-written explanation for each pair from \emph{Crowd}, along with its enhanced counterpart in \emph{C+LLM}.
This selection procedure allows us to directly compare the effect of applying LLM-based enhancement to both crowd-written and LLM-generated explanations.

\begin{figure}[t!]
    \centering
    \footnotesize
    \framebox{
    \parbox{0.45\textwidth}{
    \textbf{Context}: The band walked to the front of the stage. They began to perform. The electricity immediately went off. Everyone couldn’t see where they were. \noindent \smallskip \newline
    \textbf{Outcome}: People danced in the well lit room.\smallskip \newline
    \textbf{GPT4-generated Explanation (\emph{LLM})}: Someone in the audience had a powerful flashlight, and they used it to illuminate the room while the band continued to play acoustically. This allowed everyone to continue dancing despite the power outage. \textit{\color{purple} Comment: The explanation is detailed, but it is a less likely continuation of the context.}\smallskip \newline
    \textbf{Crowd-written Explanation (\emph{Crowd})}: Suddenly, they found the light switch. \textit{\color{purple} Comment: The explanation is likely to happen, but there is a large information gap in-between.} \smallskip \newline
    \textbf{Crowd-GPT4 Explanation (\emph{C+LLM})}: The band’s manager quickly grabbed a flashlight and located the circuit breaker, restoring power to the venue. With the electricity back on, the lights illuminated the room, allowing everyone to see and continue dancing to the band’s performance.
    \textit{\color{purple} Comment: Starting with the crowd-written explanation and refining it with an LLM results in plausible explanations that include sufficient details to connect the context and outcome.}
    }
    }
    \caption{Qualitative comparison between \llm explanations, \human explanations, and \llmhuman explanations. In \textit{\color{purple} Comments}, we make connections to the three rules in explanation writing.}
    \label{fig:qualitative_comparison}
\end{figure}
We recruit crowdworkers who provided quality explanations during data collection to provide pairwise preferences between \emph{Crowd}, \emph{C+LLM}, and \emph{LLM$^2$} explanations with \emph{LLM} explanations based on the same rules used for the explanation-writing task (Section~\ref{sec:explanations}).\footnote{Figure~\ref{fig:mturk-preference} in the appendix shows the MTurk preference evaluation template.}
Raters can select one of the two explanations as better, or can mark ties between the two as equally bad or equally good.
Table~\ref{tab:win-rate-human-gpt4} shows that \human explanations are least often preferred and \llmhuman explanations are the most preferred.
While \llm can improve via LLM-based enhancement, these explanations are still less preferred when compared to \llmhuman.
Finally, we include the Fleiss' $\kappa$ score to demonstrate the inter-annotator agreement rate between workers, where they all fall within the range from 0.40 to 0.60.\footnote{Our preference-based ranking is a four-way classification. Even though scores between 0.40 and 0.60 are considered moderate agreement for the two-class case, it is more challenging to achieve these scores in the four-class case.}
Figure~\ref{fig:qualitative_comparison} compares explanations generated by \emph{LLM}, \emph{Crowd}, and \emph{C+LLM} for an example in \story.
Table~\ref{tab:breakdown-by-likelihood} presents the non-lose rates of \llmhuman explanations against \llm explanations broken down by likelihoods.\footnote{The 100 test examples considered here only contain a significant number of outcomes with likelihoods $l=1,2$.}
\llmhuman explanations are preferable as the likelihood of outcomes are less likely.

\begin{table}[t]
    \centering
    \small
    \begin{tabular}{l c c}
    \toprule
        & \social & \story \\ \midrule
        \emph{Crowd+LLM} & 44 & 35\\
        Eql. good &33&30\\
        Eql. bad &2&1\\
        \emph{LLM} &21&34\\ \bottomrule
    \end{tabular}
    \caption{Comparing \emph{Crowd+LLM} explanations to \emph{LLM} explanations when both LLMs and crowdworkers are provided the same instrutions for producing explanations.}
    \label{tab:instruction}
\end{table}

We note that in the analysis above, we provide LLMs and crowdworkers with different instructions for producing explanations. The instructions given to crowdworkers are more detailed than those given to the LLMs. We further explore if giving LLMs the same instructions we give to humans will make LLMs perform better. We compare \emph{Crowd+LLM} and \emph{LLM} explanations and present the results in Table~\ref{tab:instruction}. We find that this instruction improves explanations on \story but harms explanations on \social. Therefore, LLMs still cannot always benefit from detailed instructions even when they include more information on what are considered good explanations.




\input{figures/lengths}

\paragraph{Quantitative Comparison of Explanations.}
We investigate several distributional differences across the four sources of explanations.
Figure~\ref{fig:lengths} shows the distribution of explanation lengths.\footnote{We use \texttt{nltk.wordpunct\_tokenize}~\cite{bird2009natural} for tokenizing explanations.}
\emph{Crowd} explanations are significantly shorter than \emph{LLM}, with an average length of \stdev{22.9}{11.3} tokens per explanation in \story and \stdev{22.0}{11.9} in \social, compared to an average of \stdev{38.2}{9.9} and \stdev{25.5}{7.1} respectively for \emph{LLM}.
However, enhancing crowd-written explanations with an LLM significantly increases their lengths over \emph{LLM}:
\emph{C+LLM} has an average explanation length of \stdev{78.0}{24.4} tokens in \story and \stdev{78.3}{23.5} in \social.
This pattern does not hold for LLM-based enhancement of LLM-generated explanations: \emph{LLM$^2$} has average lengths of \stdev{35.6}{10.8} and \stdev{25.9}{6.7} respectively, not significantly different from \emph{LLM}.
Therefore, length of the explanations produced by \emph{C+LLM} can vary significantly.

\begin{figure}[t]
\begin{center}
\footnotesize
\begin{tikzpicture}
 \begin{axis}[
    width=0.45\columnwidth,
        height=.5\columnwidth,
        xmin=0.5,xmax=5.5,
        ymin=0.6, ymax=0.9,
        xtick={1, 2, 3, 4, 5},
        bar width=12pt,
        xlabel style={yshift=0ex,},
        title=\story,
        xlabel=$n$,
        legend style={ at={(0, 1.5)}, anchor=north west, legend columns = 4},
        ylabel style={align=center, yshift=-0.2cm},
        ylabel={$n$-gram entropy}]

      \addplot[smooth, thick, dashed, red] coordinates{
        (1, 0.6597)
        (2, 0.7691)
        (3, 0.779)
        (4, 0.7717)
        (5, 0.7667)
      };
      
          \addplot[smooth, thick, red] coordinates {
          (1, 0.6737)
          (2, 0.7779)
          (3, 0.7788)
          (4, 0.7678)
          (5, 0.7612)
     }; 
     
      \addplot[smooth, thick, dashed, blue] coordinates{
      (1, 0.645)
      (2, 0.7469)
      (3, 0.7455)
      (4, 0.7316)
      (5, 0.7229)
      };

     \addplot[smooth, thick, blue] coordinates{
     (1, 0.6726)
     (2, 0.8037)
     (3, 0.83)
     (4, 0.8268)
      (5, 0.8238)
     };

    \end{axis}
\end{tikzpicture}
\begin{tikzpicture}
    
 \begin{axis}[
    width=0.45\columnwidth,
        height=.5\columnwidth,
        xmin=0,xmax=6,
        ymin=0.6, ymax=0.9,
        ytick=\empty,
        title=\social,
        xtick={1, 2, 3, 4, 5},
        bar width=12pt,
        xlabel style={yshift=0ex,},
        legend style={ at={(1.05, 1)}, anchor=north west, legend columns = 1},
        xlabel=$n$,
        ylabel style={align=center}]

      \addplot[smooth, thick, dashed, red] coordinates{
        (1, 0.6583)
        (2, 0.7821)
        (3, 0.7967)
        (4, 0.7942)
        (5, 0.7901)
      };
       \addlegendentry{\emph{LLM}}
      
          \addplot[smooth, thick, red] coordinates {
        (1, 0.6722)
        (2, 0.7922)
        (3, 0.8008)
        (4, 0.7965)
        (5, 0.7917)
     }; 
      \addlegendentry{\emph{LLM$^2$}}  
     
      \addplot[smooth, thick, dashed, blue] coordinates{
      (1, 0.6516)
      (2, 0.7829)
        (3, 0.7940)
        (4, 0.7851)
        (5, 0.7779)
      };
       \addlegendentry{\emph{Crowd}}

     \addplot[smooth, thick, blue] coordinates{
     (1, 0.6738)
     (2, 0.8252)
     (3, 0.8704)
     (4, 0.8787)
     (5, 0.8809)
     };
      \addlegendentry{\emph{C+LLM}} 

    \end{axis}
\end{tikzpicture}
\end{center}
\caption{Entropies of $n$-gram distributions in \story (left) and \social (right), computed on the development sets of each data subset. }
\label{fig:entropy}
\end{figure}
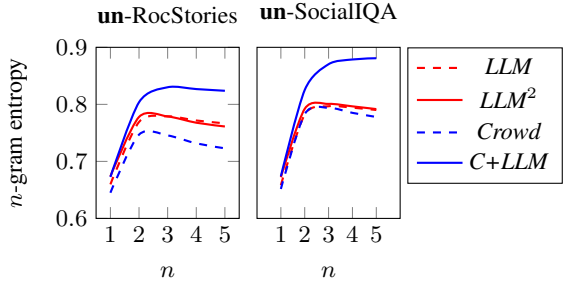

In Figure~\ref{fig:entropy}, we investigate the entropy of the distribution of $n$-grams from $n \in \{1, \dots, 5\}$ across the different sources of explanations.\footnote{As different data sources contain a different number of explanations per context-outcome pair, we compute entropy using 1,000 iterations of bootstrap sampling of one explanation per context-outcome pair in each data subset.}
We use entropy as a measure of lexical diversity~\cite{jung2023impossible}. 
We find trends similar to the analysis of explanation lengths: while \emph{Crowd} has generally lower entropy than \emph{LLM}, LLM enhancement of crowd-written explanations results in significantly higher entropy (\emph{C+LLM}), while it has no effect on LLM-generated explanations (\emph{LLM$^2$}).
Therefore, \emph{C+LLM} results in the highest lexical diversity in explanation writing.

Finally, in addition to using $n$-grams as a measure of diversity, we also perform embedding analysis to evaluate the semantic diversity of explanations written by crowdworkers and GPT-4. In particular, we compute the embedding of each \emph{crowd} explanation and each \emph{LLM} explanation\footnote{We compute the embeddings using the OpenAI ada embedding model (\texttt{text-embedding-3-large}).}, and we compute the distance between every pair of explanations for \emph{crowd} explanations and \emph{LLM} explanations, respectively. We find that the average distance between \emph{LLM} explanations is \stdev{1.26}{0.058}, while the average distance between \emph{crowd} explanations is \stdev{1.29}{0.052}, suggesting that \emph{crowd} explanations are more semantically diverse than \emph{LLM} explanations.

\section{Imitation Learning for Abductive Reasoning}
\label{sec:method}

Existing methods for abductive reasoning focus on performing supervised fine-tuning (SFT) with a static dataset~\cite{Bhagavatula2020Abductive,rudinger-etal-2020-thinking}.
Training using static demonstration data is vulnerable to exposure bias: during training, the model learns to predict the next token in an explanation conditioned on a gold-standard prefix; however, when the model generates an entirely new explanation during inference, it is conditioned on its own previously generated tokens. 
This inconsistency between training and inference procedures leads to error propagation at inference time, and a reduction in the quality of explanations. To address this issue, we experiment with several on-policy imitation learning algorithms. 



\subsection{Background: Imitation Learning}
\label{sec:imit}
In the task of abductive reasoning, a policy $\pi$ maps from the context $x$, an outcome $y$, and the prefix sequence of an explanation $z$ to a distribution over the output vocabulary.
Explanations are generated token-by-token, with the $j$th token $z_j \sim \pi(\cdot \mid x, y, z_{:j-1})$, and the entire explanation sampled from $\pi$ as $z \sim \pi (\cdot \mid x, y)$.

Let $\pi^*(\cdot)$ be an expert policy and $\pi_\theta(\cdot)$ be a learner policy with parameters $\theta$.
The objective of imitation learning is to find parameters $\theta$ that result in the learner policy assigning high probabilities to expert demonstrated explanations.

\paragraph{Behavior Cloning (BC).}
BC uses expert demonstrations $\mathcal{D}=\{(x, y, z)\}^{N}$ and a supervised learning objective that train a learner policy to maximize the probability of expert demonstrations. Existing methods of using SFT is a type of behavior cloning. 
A drawback of BC is the aforementioned exposure bias problem; as a result, 
errors are more likely to propagate during inference, where the learner fails to recover from its own mistakes, as it was never exposed to these mistakes during training.


\paragraph{Online Learning.}
To address the exposure bias problem for sequence prediction tasks, \citet{ross2011reduction} propose DAgger, where an expert policy is used at training time to provide oracle continuations to learner-generated prefixes.
The learner policy is then optimized to maximize the probability of oracle continuations, conditioned on sequence prefixes generated by the learner.
DAgger and its variants have been used in many NLP tasks, including dependency parsing~\cite{goldberg-nivre-2012-dynamic}, instruction following~\cite{anderson:17}, and language generation~\cite{lin-etal-2020-autoregressive}.


\subsection{Imitation Learning for Abductive Reasoning}
We explore two online imitation learning approaches that assume different levels of access to an expert policy, which is in our case a top-performing LLM.
First, we assume access to the expert policy at any point during training, which allows us to use it as an oracle. 
Next, we consider a setting where the expert may not be available at training time (e.g., for cost reasons), and we only have a static set of expert demonstrations.

\begin{algorithm}[t]
\footnotesize
\caption{EaO: Using expert as an oracle.}
\label{alg:dagger}
\begin{algorithmic}[1]
\State \textbf{Inputs:} Initial learner policy parameters $\theta_0$, expert policy $\pi^*(\cdot)$, dataset $\mathcal{D}=\{(x, y)\}^N$, block size $k$, initial prefix size $b$, number of training epochs $I$. 
\State $\tilde{\mathcal{D}} \gets \emptyset$ 
\For{$i=0, \dotsc, I - 1$}
    \For{$(x, y) \in \mathcal{D}$}
        \State $\tilde{z} \ \sim \pi_{\theta_i}(\cdot \ | \ x, y)$
        \State $z^* \sim \pi^*( \cdot | \ x, y, \tilde{z}_{:b})$
        \State $\tilde{\mathcal{D}} \gets \tilde{\mathcal{D}} \cup \{(x, y, \tilde{z}_{:b}z^*)\}$
    \EndFor
    \State $\theta_{i + 1} \gets \theta_i$ further optimized on $\tilde{\mathcal{D}}$ with supervised learning.
    \State $b \gets b + k$
\EndFor
\State \textbf{Returns:} Learned policy parameters $\theta_I$.
\end{algorithmic}
\end{algorithm}

\paragraph{EaO: Using \underline{e}xpert \underline{a}s an \underline{o}racle on-line.} 
Algorithm~\ref{alg:dagger} formalizes our DAgger-inspired algorithm, which we call "Expert as Oracle" (EaO). 
We train with $I$ total epochs over the training dataset $\mathcal{D} = {(x, y)}^N$. 
Throughout learning, we maintain a training dataset $\tilde{\mathcal{D}}$ containing examples of contexts and outcomes paired with explanations aggregated during each epoch.
In each epoch $i$, and for each example $(x, y)$, we use the current learner parameters $\theta_i$ to sample an explanation $\tilde{z}$.
Using a prefix $b$ of a fixed size, we then sample a continuation of $\tilde{z}_{:b}$ using the expert policy $\pi^*$.
Finally, we add an example to $\tilde{\mathcal{D}}$ that concatenates the first $b$ tokens of the learner's sample with the expert's completion. 
After aggregating examples for the epoch, we apply supervised training on examples in $\tilde{\mathcal{D}}$ to acquire updated parameters $\theta_{i + 1}$.
After each epoch, we increase the length of the prefix generated by learner policy $b$ by a fixed block size $k$. 


\label{sec:sec_with_static_expert_demos}
\paragraph{SED: Using only \underline{s}tatic \underline{e}xpert \underline{d}emonstrations.}
For the setting where we have access only to a static set of expert demonstrations, 
we propose an online learning algorithm that 
similarly aims to avoid the exposure bias problem of behavior cloning.\footnote{Full pseudocode is in Appendix~\ref{sec:sec_with_static_pseduocode}.} 

We modify the loss function of 
behavior cloning, which maximizes the probability of expert demonstration $z$, by adding two terms: (a) a term that minimizes the probability of explanations generated by the learner policy during training $\tilde z$; and (b) the KL divergence from initial policy for stabiling the training process~\cite{schulman2017ppo}. 
Formally, after sampling $\tilde z$ for each instance at each iteration from the current policy, we optimize:

\begin{small}
\begin{align}
    \label{eq:seqnll}
    \mathcal{L}(\theta) &= \frac{1}{N} \sum_{(x, y, z, \tilde{z}) \in \tilde{\mathcal{D}}}  \Big\{- \log \pi_\theta(z|x, y) + \lambda\log \pi_\theta(\tilde{z}|x, y) \nonumber \\
    &+ \beta\text{KL}\left(\pi_{\theta_0}(\cdot| x, y,z_{<t})||\pi_{\theta_i}(\cdot|x, y,z_{<t})\right) \Big\}
\end{align}
\end{small}

\begin{table*}[t]
\centering \footnotesize
\begin{tabular}{llcccccc}
\toprule
                            &            & \multicolumn{3}{c}{\story} & \multicolumn{3}{c}{\social} \\
Supervision                 & Base Model     & Win     & Tie    & Lose   & Win     & Tie     & Lose   \\ \midrule
3-shot prompting                      & GPT-3       & 13   & 20   & 67  & 33   & 13  & 54 \\ \midrule \midrule
\multirow{3}{*}{\emph{SFT} with \llm}       & GPT-2-XL    & \phantom{0}6    & 22   & 72  & \phantom{0}7    & 44    & 49  \\
                            & LLaMA-7B   & 13   & 35  & 52  & 25   & 38   & 37  \\
                            & FlanT5-XXL & 16    & 28   & 56  & 16    & 47   & 37  \\ \midrule
\multirow{3}{*}{\emph{SFT} with \llmhuman} & GPT-2-XL    & \phantom{0}6    & 26   & 64  & 13   & 32   & 55  \\
                            & LLaMA-7B   & 21   & 31   & 48  & 19   & 39    & 42  \\
                            & FlanT5-XXL & 11   & 32   & 57  & 27   & 32     & 41  \\ 
\bottomrule
\end{tabular}
\caption{Experimental comparison of GPT-3 using few-shot prompting, and \emph{SFT} with two sources of training explanations on three different base models , using pairwise preference-based evaluation on the test set of \emph{LLM}.}
\label{tab:baselines}
\end{table*}

\begin{table*}[t]
    \centering
    \footnotesize
    \begin{tabular}{c c c c}
        \toprule
        \textbf{Task} & \textbf{Uncommon?} & \textbf{Sources of Explanations} & \textbf{\# of Explanations} \\ \midrule
        $\alpha$NLG & N & Crowd workers & 76k \\
        d-NLI & N & Crowd workers & 200k \\
        Arnaout et al. & N & Variants of BERT models & N/A \\
        TODAY & Y & Crowd workers & 2.2k \\
        Collins et al. & Y & Crowd workers & 0.8k \\
        \corpus & Y & Crowd workers and GPT-4 & 41k \\ \bottomrule
    \end{tabular}
    \caption{Summary of the differences between the proposed dataset and the existing datasets.}
    \label{tab:dataset-comparison}
\end{table*}


\begin{table}[t]
\centering \footnotesize
\begin{tabular}{llccc}
\toprule
                            &        & Win   & Tie   & Lose  \\ \midrule
\multirow{3}{*}{\story}      & \emph{SFT}    & \phantom{0}6  & 22  & 72 \\
                            & \emph{SED} & 12  & 24  & 64 \\
                            & \emph{EaO} & 17  & 16  & 67 \\ \midrule
\multirow{3}{*}{\social}     & \emph{SFT}    & \phantom{0}7  & 44  & 49 \\
                            & \emph{SED} & \phantom{0}9 & 34  & 57 \\
                            & \emph{EaO} & 13 & 39 & 48 \\ \midrule \midrule
\multirow{3}{*}{$\alpha$NLG}       & \emph{SFT}    & 13 & 20  & 67 \\
                            & \emph{SED} & 14 & 23 & 63 \\
                            & \emph{EaO} & 14 & 23 & 63 \\ \midrule
\multirow{3}{*}{Sen-Making} & \emph{SFT}    & 12   & 41 & 47 \\
                            & \emph{SED} & 13  & 49 & 38 \\
                            & \emph{EaO} & 13  & 52 & 35 \\
\bottomrule
\end{tabular}
\caption{Comparison between different imitation learning methods using pairwise preference-based evaluation on the test set of \emph{LLM}.
}
\label{tab:imitation}
\end{table}

\section{Experiments}

\paragraph{Evaluation.}
We evaluate the proposed imitation learning methods with three sets of metrics.
We focus on preference-based pairwise evaluation judged by humans.\footnote{For simplicity, in this evaluation, we report equally good and equally bad as the same category (Tie).}
We report performance on the same 100 randomly-sampled examples.\footnote{We will maintain a leaderboard that provides human evaluation of these examples on model-generated explanations for two years. We will also maintain the same human annotator pool to increase reproducibility and ensure fairness.}
In Appendix~\ref{sec:evaluation}, we report two additional sets of metrics: 
(a) human judgements on seven binary questions (e.g., is the outcome more likely given the context and the explanation than given the context alone?) that evaluate different failure modes, and 
(b) a number of reference-based automatic evaluation metrics, e.g. BERTScore~\citep{Zhang2020BERTScore}. 

\paragraph{Base models.}
As baselines, we experiment with 3-shot prompting with GPT-3~\cite{brown2020language} and, following the state-of-the-art approach for commonsense abductive reasoning~\cite{khashabi-etal-2022-genie},
on several open and accessible language models: FlanT5-XXL~\cite{chung2022scaling}, LLaMA-7B~\cite{touvron2023llama}, and GPT-2-XL~\cite{radford2019language}.
To compare the benefit of different sources of training data, we perform \emph{SFT} on explanations in the training sets of \emph{LLM} (LLM-generated explanations) and \emph{C+LLM} (LLM-enhanced crowd-written explanations).
Because \emph{Crowd} (crowd-written explanations) are the least preferred subset in \corpus, we do not fine-tune on them. 
Appendix~\ref{app:experiment_details} contains additional experimental details.

\paragraph{Can imitation learning improve a given model?} We apply our proposed imitation learning algorithms, 
EaO and SED, to 
GPT-2-XL as the initial learner policy.
This is the weakest (but most computationally accessible) base language model of the three we consider for \emph{SFT}. This choice is purposeful, as our experiment intends to assess 
whether imitation learning can improve a \emph{given} LM. 
For a fair comparison, we use the same expert policy (GPT-4) for both \emph{EaO} and \emph{SED}. In addition to uncommonsense benchmarks, we report performance on two commonsense benchmarks, $\alpha$NLG~\cite{Bhagavatula2020Abductive} and Sense-making~\cite{wang-etal-2019-make} to show generalization of the methods.

\subsection{Results}
\paragraph{Baselines.}
Table~\ref{tab:baselines} shows the performance of the baseline systems. 
Unsurprisingly, explanations generated from few-shot GPT-3 are rarely preferred by crowdworkers to those GPT-4 itself generated (13\% of the time).
However, GPT-3 also underperforms the 25x smaller (but supervised fine-tuned) LLaMA-7B (48\% non-lose rate vs. GPT-4) and 16x smaller FlanT5-XLL (44\% non-lose rate vs. GPT-4). 
In addition, having \llmhuman to be supervision sometimes leads to better performance than using \llm as supervision but in other times hurts.
We hypothesize that despite \llm explanations being worse than \llmhuman explanations, they are easier for the small models to learn.
Finally, all methods but one still lose to \emph{LLM} explanations, indicating that \emph{SFT} alone is insufficient.

\paragraph{Imitation Learning.}
Table~\ref{tab:imitation} shows the performance comparing \emph{SFT} with the two imitation learning methods, \emph{SED} and \emph{EaO}, on four datasets when using GPT-2-XL as the base moddel.
On both \corpus and commonsense benchmarks, \emph{SED} and \emph{EaO} show strong improvements against \emph{SFT} by reducing the losing rate to \emph{LLM} explanations or by increasing the win rates.
Except for $\alpha$NLG, \emph{EaO}, which trains using expert online corrections to learner-generated sequence prefixes, shows more promise than \emph{SED} on most of the datasets.
However, \emph{SED}, which is no more costly than \emph{SFT}, can significantly improve the performance of the weak-but-accessible base model GPT-2-XL on both commonsense and uncommonsense reasoning except on \social.

\section{Related Work}
$\alpha$NLG~\cite{Bhagavatula2020Abductive} is the most closely related task to \corpus: both require generating explanations to bridge contexts and outcomes (except $\alpha$NLG focuses on common, everyday scenarios).
d-NLI~\cite{rudinger-etal-2020-thinking} consider a related task of generating an explanation explanation that weakens an outcome.
Additional works cover methods for generating explanations, e.g., \citet{du-etal-2022-e}, \citet{zhou-etal-2021-probing-commonsense}, \citet{wang-etal-2019-make}, \citet{zhang-etal-2020-winowhy}, inter alia.

Reasoning about uncommon but possible scenarios has been studied in other settings.
\citet{arnaout2022uncommonsense} propose a method for identifying informative negations about everyday concepts in large-scale commonsense knowledge bases.
\citet{tang-etal-2023-less} present a decoding method for producing less plausible explanations for everyday events.
\citet{collins2022structured} create a small-scale benchmark containing about 800 curated uncommon statements, along with explanations making sense of these statements.
\corpus differs in structure and focus from these prior works.
Finally, TODAY \cite{feng-etal-2023-generic} proposes a temporal reasoning task to study the order of two events. Atypical order of two events could be uncommon, and justifying the order is uncommonsense reasoning.
Because \corpus is not built from reversing the order of temporal events, it encompasses a different set of uncommon situations, including social reasoning, cultural reasoning, and physical reasoning. With each situation, \corpus also contains more than one explanation, collected from both crowd workers and GPT-4.
We summarize the differences between \corpus and existing datasets in Figure \ref{tab:dataset-comparison}.

Finally, uncommonsense reasoning is closely related to defeasible reasoning~\cite{rudinger-etal-2020-thinking,madaan-etal-2021-give,madaan-etal-2021-think}. Both defeasible reasoning and reasoning about uncommon situations are, given context $x$ and outcome $y$, finding an explanation $z$ that changes the original likelihood $p(y|x)$ by adding z: $p(y|x, z)$.
However, we note that feasible reasoning itself does not place any constraint on $p(y|x)$. Reasoning about uncommon situations falls on the long-tail distribution of defeasible reasoning as it focuses on the cases where $p(y|x)$ is very small. 

\section{Conclusion}
We propose a new task, uncommonsense abductive reasoning, designed to assess the ability of NLP systems to reason about uncommon scenarios in abductive reasoning tasks.
We explore two imitation learning methods to improve accessible language models on uncommonsense abductive reasoning. 
Experiments show that access to expert behavior, particularly when using the expert as an oracle in online training, significantly improves the explanation quality of smaller models.

\section*{Limitations}
While our dataset offers advantages over existing sources, we acknowledge the following limitations.
First, our dataset may suffer from social biases in the data collection process, and the labeling process may contain errors and inconsistencies. Despite best efforts to ensure high-quality annotations, occasional human errors are possible.
Additionally, our dataset only contains uncommon situations in English and thus lack of language diversity.
Finally, our main preference-based evaluation relies on human evaluators, which can be less producible and costly.
There is thus a large room for improvement for more effective and affordable evaluation methods.

\section*{Ethics Statement}
This work aims to advance NLP and commonsense reasoning by introducing a new benchmark, \corpus, which investigates abductive reasoning about uncommon events. It is important to study these uncommon situations as they provide valuable insights into the proper functioning of AI systems in real-world, unpredictable circumstances. However, we emphasize the need to ensure that the generation of natural language explanations follows ethical guidelines and respects privacy, diversity, and fairness. We are committed to maintaining transparency and sharing the code and data, fostering open collaboration to address potential ethical concerns and promote the responsible advancement of AI technologies.

\section*{Acknowledgments}
We thank the anonymous reviewers for their feedback to help us improve the paper. We thank Xiang Ren, Xinyan Yu, and Sasha Rush for numerous helpful discussions. This work was partially supported by an AI2 Young Investigator Grant.


\bibliography{anthology,custom}

\begin{thebibliography}{33}
\expandafter\ifx\csname natexlab\endcsname\relax\def\natexlab#1{#1}\fi

\bibitem[{Anderson et~al.(2017)Anderson, Wu, Teney, Bruce, Johnson, S{\"u}nderhauf, Reid, Gould, and van~den Hengel}]{anderson:17}
Peter Anderson, Qi~Wu, Damien Teney, Jake Bruce, Mark Johnson, Niko S{\"u}nderhauf, Ian~D. Reid, Stephen Gould, and Anton van~den Hengel. 2017.
\newblock Vision-and-language navigation: Interpreting visually-grounded navigation instructions in real environments.
\newblock \emph{2018 IEEE/CVF Conference on Computer Vision and Pattern Recognition}, pages 3674--3683.

\bibitem[{Arnaout et~al.(2022)Arnaout, Razniewski, Weikum, and Pan}]{arnaout2022uncommonsense}
Hiba Arnaout, Simon Razniewski, Gerhard Weikum, and Jeff~Z Pan. 2022.
\newblock Un{C}ommon{S}ense: Informative negative knowledge about everyday concepts.
\newblock In \emph{Proceedings of the 31st ACM International Conference on Information \& Knowledge Management}, pages 37--46.

\bibitem[{Bhagavatula et~al.(2020)Bhagavatula, Bras, Malaviya, Sakaguchi, Holtzman, Rashkin, Downey, tau Yih, and Choi}]{Bhagavatula2020Abductive}
Chandra Bhagavatula, Ronan~Le Bras, Chaitanya Malaviya, Keisuke Sakaguchi, Ari Holtzman, Hannah Rashkin, Doug Downey, Wen tau Yih, and Yejin Choi. 2020.
\newblock \href {https://openreview.net/forum?id=Byg1v1HKDB} {Abductive commonsense reasoning}.
\newblock In \emph{International Conference on Learning Representations}.

\bibitem[{Bird et~al.(2009)Bird, Klein, and Loper}]{bird2009natural}
Steven Bird, Ewan Klein, and Edward Loper. 2009.
\newblock \emph{Natural language processing with {P}ython: analyzing text with the natural language toolkit}.
\newblock " O'Reilly Media, Inc.".

\bibitem[{Brown et~al.(2020)Brown, Mann, Ryder, Subbiah, Kaplan, Dhariwal, Neelakantan, Shyam, Sastry, Askell et~al.}]{brown2020language}
Tom Brown, Benjamin Mann, Nick Ryder, Melanie Subbiah, Jared~D Kaplan, Prafulla Dhariwal, Arvind Neelakantan, Pranav Shyam, Girish Sastry, Amanda Askell, et~al. 2020.
\newblock Language models are few-shot learners.
\newblock \emph{Advances in neural information processing systems}, 33:1877--1901.

\bibitem[{Chung et~al.(2022)Chung, Hou, Longpre, Zoph, Tay, Fedus, Li, Wang, Dehghani, Brahma et~al.}]{chung2022scaling}
Hyung~Won Chung, Le~Hou, Shayne Longpre, Barret Zoph, Yi~Tay, William Fedus, Eric Li, Xuezhi Wang, Mostafa Dehghani, Siddhartha Brahma, et~al. 2022.
\newblock Scaling instruction-finetuned language models.
\newblock \emph{arXiv preprint arXiv:2210.11416}.

\bibitem[{Collins et~al.(2022)Collins, Wong, Feng, Wei, and Tenenbaum}]{collins2022structured}
Katherine~M Collins, Catherine Wong, Jiahai Feng, Megan Wei, and Josh Tenenbaum. 2022.
\newblock Structured, flexible, and robust: benchmarking and improving large language models towards more human-like behavior in out-of-distribution reasoning tasks.
\newblock In \emph{Proceedings of the Annual Meeting of the Cognitive Science Society}, volume~44.

\bibitem[{Du et~al.(2022)Du, Ding, Xiong, Liu, and Qin}]{du-etal-2022-e}
Li~Du, Xiao Ding, Kai Xiong, Ting Liu, and Bing Qin. 2022.
\newblock \href {https://doi.org/10.18653/v1/2022.acl-long.33} {e-{CARE}: a new dataset for exploring explainable causal reasoning}.
\newblock In \emph{Proceedings of the 60th Annual Meeting of the Association for Computational Linguistics (Volume 1: Long Papers)}, pages 432--446, Dublin, Ireland. Association for Computational Linguistics.

\bibitem[{Feng et~al.(2023)Feng, Zhou, Wang, Jin, and Roth}]{feng-etal-2023-generic}
Yu~Feng, Ben Zhou, Haoyu Wang, Helen Jin, and Dan Roth. 2023.
\newblock \href {https://doi.org/10.18653/v1/2023.acl-long.671} {Generic temporal reasoning with differential analysis and explanation}.
\newblock In \emph{Proceedings of the 61st Annual Meeting of the Association for Computational Linguistics (Volume 1: Long Papers)}, pages 12013--12029, Toronto, Canada. Association for Computational Linguistics.

\bibitem[{Goldberg and Nivre(2012)}]{goldberg-nivre-2012-dynamic}
Yoav Goldberg and Joakim Nivre. 2012.
\newblock \href {https://aclanthology.org/C12-1059} {A dynamic oracle for arc-eager dependency parsing}.
\newblock In \emph{Proceedings of {COLING} 2012}, pages 959--976, Mumbai, India. The COLING 2012 Organizing Committee.

\bibitem[{Jung et~al.(2023)Jung, West, Jiang, Brahman, Lu, Fisher, Sorensen, and Choi}]{jung2023impossible}
Jaehun Jung, Peter West, Liwei Jiang, Faeze Brahman, Ximing Lu, Jillian Fisher, Taylor Sorensen, and Yejin Choi. 2023.
\newblock \href {http://arxiv.org/abs/2305.16635} {Impossible distillation: from low-quality model to high-quality dataset and model for summarization and paraphrasing}.

\bibitem[{Khashabi et~al.(2022)Khashabi, Stanovsky, Bragg, Lourie, Kasai, Choi, Smith, and Weld}]{khashabi-etal-2022-genie}
Daniel Khashabi, Gabriel Stanovsky, Jonathan Bragg, Nicholas Lourie, Jungo Kasai, Yejin Choi, Noah~A. Smith, and Daniel Weld. 2022.
\newblock \href {https://doi.org/10.18653/v1/2022.emnlp-main.787} {{GENIE}: Toward reproducible and standardized human evaluation for text generation}.
\newblock In \emph{Proceedings of the 2022 Conference on Empirical Methods in Natural Language Processing}, pages 11444--11458, Abu Dhabi, United Arab Emirates. Association for Computational Linguistics.

\bibitem[{Lin et~al.(2020{\natexlab{a}})Lin, Wohlwend, Chen, and Lei}]{lin-etal-2020-autoregressive}
Alexander Lin, Jeremy Wohlwend, Howard Chen, and Tao Lei. 2020{\natexlab{a}}.
\newblock \href {https://doi.org/10.18653/v1/2020.emnlp-main.494} {Autoregressive knowledge distillation through imitation learning}.
\newblock In \emph{Proceedings of the 2020 Conference on Empirical Methods in Natural Language Processing (EMNLP)}, pages 6121--6133, Online. Association for Computational Linguistics.

\bibitem[{Lin et~al.(2020{\natexlab{b}})Lin, Zhou, Shen, Zhou, Bhagavatula, Choi, and Ren}]{lin-etal-2020-commongen}
Bill~Yuchen Lin, Wangchunshu Zhou, Ming Shen, Pei Zhou, Chandra Bhagavatula, Yejin Choi, and Xiang Ren. 2020{\natexlab{b}}.
\newblock \href {https://doi.org/10.18653/v1/2020.findings-emnlp.165} {{C}ommon{G}en: A constrained text generation challenge for generative commonsense reasoning}.
\newblock In \emph{Findings of the Association for Computational Linguistics: EMNLP 2020}, pages 1823--1840, Online. Association for Computational Linguistics.

\bibitem[{Madaan et~al.(2021{\natexlab{a}})Madaan, Rajagopal, Tandon, Yang, and Hovy}]{madaan-etal-2021-give}
Aman Madaan, Dheeraj Rajagopal, Niket Tandon, Yiming Yang, and Eduard Hovy. 2021{\natexlab{a}}.
\newblock \href {https://doi.org/10.18653/v1/2021.findings-acl.456} {Could you give me a hint ? generating inference graphs for defeasible reasoning}.
\newblock In \emph{Findings of the Association for Computational Linguistics: ACL-IJCNLP 2021}, pages 5138--5147, Online. Association for Computational Linguistics.

\bibitem[{Madaan et~al.(2021{\natexlab{b}})Madaan, Tandon, Rajagopal, Clark, Yang, and Hovy}]{madaan-etal-2021-think}
Aman Madaan, Niket Tandon, Dheeraj Rajagopal, Peter Clark, Yiming Yang, and Eduard Hovy. 2021{\natexlab{b}}.
\newblock \href {https://doi.org/10.18653/v1/2021.emnlp-main.508} {Think about it! improving defeasible reasoning by first modeling the question scenario.}
\newblock In \emph{Proceedings of the 2021 Conference on Empirical Methods in Natural Language Processing}, pages 6291--6310, Online and Punta Cana, Dominican Republic. Association for Computational Linguistics.

\bibitem[{Mostafazadeh et~al.(2016{\natexlab{a}})Mostafazadeh, Chambers, He, Parikh, Batra, Vanderwende, Kohli, and Allen}]{mostafazadeh-etal-2016-corpus}
Nasrin Mostafazadeh, Nathanael Chambers, Xiaodong He, Devi Parikh, Dhruv Batra, Lucy Vanderwende, Pushmeet Kohli, and James Allen. 2016{\natexlab{a}}.
\newblock \href {https://doi.org/10.18653/v1/N16-1098} {A corpus and cloze evaluation for deeper understanding of commonsense stories}.
\newblock In \emph{Proceedings of the 2016 Conference of the North {A}merican Chapter of the Association for Computational Linguistics: Human Language Technologies}, pages 839--849, San Diego, California. Association for Computational Linguistics.

\bibitem[{Mostafazadeh et~al.(2016{\natexlab{b}})Mostafazadeh, Grealish, Chambers, Allen, and Vanderwende}]{mostafazadeh-etal-2016-caters}
Nasrin Mostafazadeh, Alyson Grealish, Nathanael Chambers, James Allen, and Lucy Vanderwende. 2016{\natexlab{b}}.
\newblock \href {https://doi.org/10.18653/v1/W16-1007} {{C}a{T}e{RS}: Causal and temporal relation scheme for semantic annotation of event structures}.
\newblock In \emph{Proceedings of the Fourth Workshop on Events}, pages 51--61, San Diego, California. Association for Computational Linguistics.

\bibitem[{OpenAI(2023)}]{openai2023gpt4}
OpenAI. 2023.
\newblock \href {http://arxiv.org/abs/2303.08774} {{GPT}-4 technical report}.

\bibitem[{Radford et~al.(2019)Radford, Wu, Child, Luan, Amodei, Sutskever et~al.}]{radford2019language}
Alec Radford, Jeffrey Wu, Rewon Child, David Luan, Dario Amodei, Ilya Sutskever, et~al. 2019.
\newblock Language models are unsupervised multitask learners.
\newblock \emph{OpenAI blog}, 1(8):9.

\bibitem[{Ross et~al.(2011)Ross, Gordon, and Bagnell}]{ross2011reduction}
St{\'e}phane Ross, Geoffrey Gordon, and Drew Bagnell. 2011.
\newblock A reduction of imitation learning and structured prediction to no-regret online learning.
\newblock In \emph{Proceedings of the fourteenth international conference on artificial intelligence and statistics}, pages 627--635. JMLR Workshop and Conference Proceedings.

\bibitem[{Rudinger et~al.(2020)Rudinger, Shwartz, Hwang, Bhagavatula, Forbes, Le~Bras, Smith, and Choi}]{rudinger-etal-2020-thinking}
Rachel Rudinger, Vered Shwartz, Jena~D. Hwang, Chandra Bhagavatula, Maxwell Forbes, Ronan Le~Bras, Noah~A. Smith, and Yejin Choi. 2020.
\newblock \href {https://doi.org/10.18653/v1/2020.findings-emnlp.418} {Thinking like a skeptic: Defeasible inference in natural language}.
\newblock In \emph{Findings of the Association for Computational Linguistics: EMNLP 2020}, pages 4661--4675, Online. Association for Computational Linguistics.

\bibitem[{Sap et~al.(2019)Sap, Rashkin, Chen, Le~Bras, and Choi}]{sap-etal-2019-social}
Maarten Sap, Hannah Rashkin, Derek Chen, Ronan Le~Bras, and Yejin Choi. 2019.
\newblock \href {https://doi.org/10.18653/v1/D19-1454} {Social {IQ}a: Commonsense reasoning about social interactions}.
\newblock In \emph{Proceedings of the 2019 Conference on Empirical Methods in Natural Language Processing and the 9th International Joint Conference on Natural Language Processing (EMNLP-IJCNLP)}, pages 4463--4473, Hong Kong, China. Association for Computational Linguistics.

\bibitem[{Schulman et~al.(2017)Schulman, Wolski, Dhariwal, Radford, and Klimov}]{schulman2017ppo}
John Schulman, Filip Wolski, Prafulla Dhariwal, Alec Radford, and Oleg Klimov. 2017.
\newblock \href {http://dblp.uni-trier.de/db/journals/corr/corr1707.html#SchulmanWDRK17} {Proximal policy optimization algorithms.}
\newblock \emph{CoRR}, abs/1707.06347.

\bibitem[{Talmor et~al.(2019)Talmor, Herzig, Lourie, and Berant}]{talmor-etal-2019-commonsenseqa}
Alon Talmor, Jonathan Herzig, Nicholas Lourie, and Jonathan Berant. 2019.
\newblock \href {https://doi.org/10.18653/v1/N19-1421} {{C}ommonsense{QA}: A question answering challenge targeting commonsense knowledge}.
\newblock In \emph{Proceedings of the 2019 Conference of the North {A}merican Chapter of the Association for Computational Linguistics: Human Language Technologies, Volume 1 (Long and Short Papers)}, pages 4149--4158, Minneapolis, Minnesota. Association for Computational Linguistics.

\bibitem[{Tang et~al.(2023)Tang, Peng, Wang, Ding, Durrett, and Rousseau}]{tang-etal-2023-less}
Liyan Tang, Yifan Peng, Yanshan Wang, Ying Ding, Greg Durrett, and Justin Rousseau. 2023.
\newblock \href {https://doi.org/10.18653/v1/2023.findings-acl.794} {Less likely brainstorming: Using language models to generate alternative hypotheses}.
\newblock In \emph{Findings of the Association for Computational Linguistics: ACL 2023}, pages 12532--12555, Toronto, Canada. Association for Computational Linguistics.

\bibitem[{Touvron et~al.(2023)Touvron, Lavril, Izacard, Martinet, Lachaux, Lacroix, Rozi{\`e}re, Goyal, Hambro, Azhar et~al.}]{touvron2023llama}
Hugo Touvron, Thibaut Lavril, Gautier Izacard, Xavier Martinet, Marie-Anne Lachaux, Timoth{\'e}e Lacroix, Baptiste Rozi{\`e}re, Naman Goyal, Eric Hambro, Faisal Azhar, et~al. 2023.
\newblock Llama: Open and efficient foundation language models.
\newblock \emph{arXiv preprint arXiv:2302.13971}.

\bibitem[{Wang et~al.(2019)Wang, Liang, Zhang, Li, and Gao}]{wang-etal-2019-make}
Cunxiang Wang, Shuailong Liang, Yue Zhang, Xiaonan Li, and Tian Gao. 2019.
\newblock \href {https://doi.org/10.18653/v1/P19-1393} {Does it make sense? and why? a pilot study for sense making and explanation}.
\newblock In \emph{Proceedings of the 57th Annual Meeting of the Association for Computational Linguistics}, pages 4020--4026, Florence, Italy. Association for Computational Linguistics.

\bibitem[{Weidinger et~al.(2022)Weidinger, Uesato, Rauh, Griffin, Huang, Mellor, Glaese, Cheng, Balle, Kasirzadeh et~al.}]{weidinger2022taxonomy}
Laura Weidinger, Jonathan Uesato, Maribeth Rauh, Conor Griffin, Po-Sen Huang, John Mellor, Amelia Glaese, Myra Cheng, Borja Balle, Atoosa Kasirzadeh, et~al. 2022.
\newblock Taxonomy of risks posed by language models.
\newblock In \emph{Proceedings of the 2022 ACM Conference on Fairness, Accountability, and Transparency}, pages 214--229.

\bibitem[{Wolf et~al.(2020)Wolf, Debut, Sanh, Chaumond, Delangue, Moi, Cistac, Rault, Louf, Funtowicz, Davison, Shleifer, von Platen, Ma, Jernite, Plu, Xu, Le~Scao, Gugger, Drame, Lhoest, and Rush}]{wolf-etal-2020-transformers}
Thomas Wolf, Lysandre Debut, Victor Sanh, Julien Chaumond, Clement Delangue, Anthony Moi, Pierric Cistac, Tim Rault, Remi Louf, Morgan Funtowicz, Joe Davison, Sam Shleifer, Patrick von Platen, Clara Ma, Yacine Jernite, Julien Plu, Canwen Xu, Teven Le~Scao, Sylvain Gugger, Mariama Drame, Quentin Lhoest, and Alexander Rush. 2020.
\newblock \href {https://doi.org/10.18653/v1/2020.emnlp-demos.6} {Transformers: State-of-the-art natural language processing}.
\newblock In \emph{Proceedings of the 2020 Conference on Empirical Methods in Natural Language Processing: System Demonstrations}, pages 38--45, Online. Association for Computational Linguistics.

\bibitem[{Zhang et~al.(2020{\natexlab{a}})Zhang, Zhao, and Song}]{zhang-etal-2020-winowhy}
Hongming Zhang, Xinran Zhao, and Yangqiu Song. 2020{\natexlab{a}}.
\newblock \href {https://doi.org/10.18653/v1/2020.acl-main.508} {{W}ino{W}hy: A deep diagnosis of essential commonsense knowledge for answering {W}inograd schema challenge}.
\newblock In \emph{Proceedings of the 58th Annual Meeting of the Association for Computational Linguistics}, pages 5736--5745, Online. Association for Computational Linguistics.

\bibitem[{Zhang et~al.(2020{\natexlab{b}})Zhang, Kishore, Wu, Weinberger, and Artzi}]{Zhang2020BERTScore}
Tianyi Zhang, Varsha Kishore, Felix Wu, Kilian~Q. Weinberger, and Yoav Artzi. 2020{\natexlab{b}}.
\newblock \href {https://openreview.net/forum?id=SkeHuCVFDr} {{BERTScore}: Evaluating text generation with {BERT}}.
\newblock In \emph{International Conference on Learning Representations}.

\bibitem[{Zhou et~al.(2021)Zhou, Jandaghi, Cho, Lin, Pujara, and Ren}]{zhou-etal-2021-probing-commonsense}
Pei Zhou, Pegah Jandaghi, Hyundong Cho, Bill~Yuchen Lin, Jay Pujara, and Xiang Ren. 2021.
\newblock \href {https://doi.org/10.18653/v1/2021.findings-emnlp.349} {Probing commonsense explanation in dialogue response generation}.
\newblock In \emph{Findings of the Association for Computational Linguistics: EMNLP 2021}, pages 4132--4146, Punta Cana, Dominican Republic. Association for Computational Linguistics.

\end{thebibliography}
\bibliographystyle{acl_natbib}

\appendix

\begin{table*}[ht!]
\footnotesize
\centering
\begin{tabular}{p{0.001\textwidth} p{0.5\textwidth} p{0.38\textwidth}}
 \toprule
  $l$ &
  Context &
  Outcome \\ \midrule
  4 &
  Kate and Greg went to a little candy shop together. They looked around at their options and made their choice. They went up to the cashier and said what they wanted. The cashier, with unwashed hands, bagged the candy without gloves. &
  Kate and Greg licked the candy gleefully. \\ \midrule
  3 &
  I went to the post office yesterday. It took a while to get there since it's on the other side of town. Once I got there I mailed my letters and headed home. It's always easier to get home than to get somewhere. &
  I could not find my way back from the post office. \\ \midrule
  2 &
  My niece just got engaged. She is Chinese and her fiance is Caucasian. Her parents had them over for a home cooked meal. The fiance got nausea from the unfamiliar dishes and had to leave. &
  My niece was thrilled that her fiance was sick. \\ \midrule
  1 &
  Josh woke up early to get ready for the hike he had been planning. After a shower, he made sure all his supplies were packed. He left his house and drove to the park where he was going hiking. Because it was early in the day Josh had the trail mostly to himself. &
  Josh loathed the outdoors. \\ \midrule
  4 &
  Jordan finished their test so fast and still got an A plus as always. &
  Other students will be jealous. \\ \midrule
  3 &
  Skylar gave Robin the permission to eat cake after Robin caused some trouble. &
  Robin will want to refuse to eat the cake. \\ \midrule
  2 &
  Austin brought tears to Tracy's eyes when he brought her flowers. &
  Austin will be hated. \\ \midrule
  1 &
  Carson threw beer in Kendall's face during a heated argument with her. &
  Carson will receive a medal for their behavior.\\
  \bottomrule
\end{tabular}
\caption{Example outcomes of different likelihood scores $l \in \{4,3,2,1\}$.}
\label{tab:qualitative-outcome}
\end{table*}
\section{Qualitative Analysis of Outcomes}
Table~\ref{tab:qualitative-outcome} presents example outcomes of different likelihood scores.

\section{Processing Outcomes in SocialIQA}
\label{sec:processing_socialiqa}
 We use three types of questions: what will X want to do next, what will happen to X, and how would you describe X. We do the following steps to construct the outcome.
\begin{enumerate}
    \item We remove the correct answer choice, and we are left with two incorrect answer choices.
    \item We feed GPT3 (text-davinci-03) ``\{context\} \{question\} \{answer\}'' and compute the answer probability $p$ (answer | context, question) and choose the answer that has the lower probability.
    \item We prompt ChatGPT to combine the question and the answer to be the outcome, in the six-shot setting. When we receive a response from ChatGPT, we check whether the original answer is in the output, if it doesn't contain the answer, we send the same prompt to GPT-4. If GPT-4 still fails, we mark the example and manually combine the question and the answer. Refer to \ref{fig:combine-question-answer-socialiqa} for the combining prompting template.
\end{enumerate}
Because SocialIQA contains many invalid answer choices, the combining step often fails (e.g., the question is ``what will person X do next'', and the answer is ``sad''), we rely on ChatGPT to detect such cases. We throw out the examples when ChatGPT refuses to do the combination.
\begin{table*}[t]
\centering
\footnotesize
\begin{tabular}{llccccccc}
\toprule
&&\multicolumn{3}{c}{Consistency}&\multicolumn{2}{c}{Relevance}&\multicolumn{2}{c}{Plausibility}\\

\textbf{Supervision}&\textbf{Model} & $x$ & $y$ & $z$ & $x$ & $y$ & $z$ & $y$ \\ \midrule
\multirow{2}{*}{3-shot prompting}& GPT-3 & 76 & 92 & 97 & 100 & 97 & 74 & 75 \\
 &GPT-4 & 94 & 91 & 99 & 98 & 95 & 90 & 85 \\ \midrule
\multirow{3}{*}{\emph{SFT with LLM}} & LlaMA-7B & 89 & 92 & 98 & 98 & 95 & 89 & 80 \\
 & FlanT5-XXL & 80 & 93 & 94 & 96 & 93 & 74 & 58 \\
 & GPT-2-XL & 88 & 92 & 92 & 97 & 94 & 80 & 83 \\ \midrule
\emph{SED with LLM} & GPT-2-XL & 78 & 87 & 90 & 94 & 94 & 66 & 77 \\\midrule
\emph{EoA with LLM} & GPT-2-XL & 97 & 94 & 94 & 100 & 94 & 88 & 86 \\
\bottomrule
\end{tabular}
\caption{Fine-grained human evaluation on \story.}
\label{tab:rocstories-fine-grained}
\end{table*}

\begin{table*}[t]
\centering
\footnotesize
\begin{tabular}{llccccccc}
\toprule
&&\multicolumn{3}{c}{Consistency}&\multicolumn{2}{c}{Relevance}&\multicolumn{2}{c}{Plausibility}\\

\textbf{Supervision}&\textbf{Model} & $x$ & $y$ & $z$ & $x$ & $y$ & $z$ & $y$ \\ \midrule
\multirow{2}{*}{Few-shot prompting}& GPT-3 & 93 & 92 & 99 & 100 & 94 & 93 & 84 \\
 &GPT-4 & 98 & 92 & 100 & 99 & 97 & 95 & 90 \\ \midrule
\multirow{3}{*}{\emph{SF with LLM}} & LlaMA-7B & 93 & 94 & 100 & 99 & 97 & 90 & 88 \\
 & FlanT5-XXL & 94 & 91 & 97 & 96 & 96 & 88 & 80 \\
 & GPT-2-XL & 96 & 95 & 97 & 98 & 98 & 91 & 91 \\ \midrule
\emph{SED with LLM} & GPT-2-XL & 94 & 84 & 97 & 97 & 85 & 92 & 73 \\\midrule
\emph{EoA with LLM} & GPT-2-XL & 91 & 95 & 97 & 97 & 94 & 87 & 88 \\
\bottomrule
\end{tabular}
\caption{Fine-grained human evaluation on \social.}
\label{tab:socialiqa-fine-grained}
\end{table*}

\begin{table*}[t]
\centering
\footnotesize
\begin{tabular}{llccccccc}
\toprule
&&\multicolumn{3}{c}{Consistency}&\multicolumn{2}{c}{Relevance}&\multicolumn{2}{c}{Plausibility}\\

\textbf{Supervision}&\textbf{Model} & $x$ & $y$ & $z$ & $x$ & $y$ & $z$ & $y$ \\ \midrule
\multirow{2}{*}{Few-shot prompting}& GPT-3 & 100 & 97 & 100 & 99 & 96 & 98 & 94 \\
 &GPT-4 & 99 & 98 & 99 & 99 & 98 & 99 & 97 \\ \midrule
\multirow{3}{*}{\emph{SFT with LLM}} & LlaMA-7B & 99 & 97 & 99 & 95 & 97 &98& 91 \\
 & FlanT5-XXL & 95 & 92 & 95 & 93 & 81 & 96 & 85 \\
 & GPT-2-XL & 96 & 92 & 98 & 93 & 97 & 94 & 85 \\ \midrule
\emph{SED with LLM} & GPT-2-XL & 97 & 91 & 99 & 95 & 93 & 96 & 84 \\\midrule
\emph{EoA with LLM} & GPT-2-XL & 97 & 95 & 98 & 97 & 97 & 98 & 90 \\
\bottomrule
\end{tabular}
\caption{Fine-grained human evaluation on $\alpha$NLG.}
\label{tab:anlg-fine-grained}
\end{table*}

\begin{table*}[t]
\centering
\footnotesize
\begin{tabular}{llccc}
\toprule
& & \multicolumn{1}{c}{Consistency}&\multicolumn{1}{c}{Relevance}&\multicolumn{1}{c}{Plausibility}\\
\textbf{Supervision}&\textbf{Model} & \multicolumn{1}{c}{$y$}&\multicolumn{1}{c}{$y$}&\multicolumn{1}{c}{$y$}\\

\midrule
\multirow{2}{*}{Few-shot prompting}& GPT-3 & 100 & 100 & 93 \\
 &GPT-4 & 100 & 100 & 99 \\ \midrule
\multirow{3}{*}{\emph{SFT with LLM}} & LlaMA-7B & 91 & 95 & 86 \\
 & FlanT5-XXL & 85 & 98 & 84 \\
 & GPT-2-XL & 86 & 97 & 85 \\ \midrule
\emph{SED with LLM} & GPT-2-XL & 92 & 98 & 91 \\\midrule
\emph{EoA with LLM} & GPT-2-XL & 87 & 95 & 83 \\
\bottomrule
\end{tabular}
\caption{Fine-grained human evaluation on Sen-making.}
\label{tab:sen-making-fine-grained}
\end{table*}

\section{More Evaluation}
\label{sec:evaluation}
We include additional automatic and human evaluation results on baseline models and our proposed imitation learning methods, \emph{SED} and \emph{EaO}.
The additional human evaluation is is a set of seven human evaluation questions that target different failure modes of generated explanations:
\begin{enumerate}
    \item Is the explanation relevant to the context? (denoted as relevance $x$)
    \item Is the explanation relevant to the outcome? (denoted as relevance $y$)
    \item Is the explanation not self-contradictory? (denoted as consistency $z$)
    \item Is the explanation not contradictory to the context? (denoted as consistency $x$)
    \item Is the explanation not contradictory to the outcome? (denoted as consistency $y$)
    \item Is it possible that explanation might occur (given the context)? (denoted as plausibility $z$)
    \item Is the outcome more likely given the context and the explanation than given the context alone? (plausibility $y$)
\end{enumerate}
The results are presented in Table~\ref{tab:rocstories-fine-grained} for \story, Table~\ref{tab:socialiqa-fine-grained} for \social, Table~\ref{tab:anlg-fine-grained} for $\alpha$NLG, and Table~\ref{tab:sen-making-fine-grained} for Sen-Making.

We also compute BERTScore, ROUGE-L, METEOR, SacreBLEU, and BLEURT for each method and report the results in Table~\ref{tab:story-auto} for \story, Table~\ref{tab:social-auto} for \social, Table~\ref{tab:anlg-auto} for $\alpha$NLG, and Table~\ref{tab:sen-auto} for Sen-Making.

\begin{table*}[t]
\centering
\footnotesize
\begin{tabular}{llccccc}
\toprule
\textbf{Supervision}&\textbf{Model} & \multicolumn{1}{c}{BERTScore} & \multicolumn{1}{c}{ROUGE} & \multicolumn{1}{c}{METEOR} & \multicolumn{1}{c}{SacreBLEU} & \multicolumn{1}{c}{BLEURT} \\ \midrule
3-shot prompting &GPT-4       & 90.79 & 30.43 & 29.79 & 5.16 & -21.74 \\ \midrule
\emph{SFT with LLM}& GPT2-XL & 90.01 & 26.67 & 24.17 & 3.44 & -35.22 \\ \midrule
\emph{SED with LLM}& GPT2-XL & 89.76 & 26.08 & 22.64 & 2.94 & -40.04 \\ \midrule
\emph{EaO with LLM}& GPT2-XL & 89.91 & 26.79 & 25.72 & 3.94 & -30.32 \\
\bottomrule
\end{tabular}
\caption{Automatic evaluation on \story.}
\label{tab:story-auto}
\end{table*}

\begin{table*}[t]
\centering
\footnotesize
\begin{tabular}{llccccc}
\toprule
\textbf{Supervision}&\textbf{Model} & \multicolumn{1}{c}{BERTScore} & \multicolumn{1}{c}{ROUGE} & \multicolumn{1}{c}{METEOR} & \multicolumn{1}{c}{SacreBLEU} & \multicolumn{1}{c}{BLEURT} \\ \midrule
3-shot prompting & GPT-4 & 90.79 & 30.43 & 29.79 & 5.16 & -21.74 \\ \midrule
\emph{SFT with LLM}& GPT2-XL & 90.01 & 26.67 & 24.17 & 3.44 & -35.22 \\ \midrule
\emph{SED with LLM}& GPT2-XL & 89.76 & 26.08 & 22.64 & 2.94 & -40.04 \\ \midrule
\emph{EaO with LLM}& GPT2-XL & 89.91 & 26.79 & 25.72 & 3.94 & -30.32 \\
\bottomrule
\end{tabular}
\caption{Automatic evaluation on \social.}
\label{tab:social-auto}
\end{table*}

\begin{table*}[t]
\centering
\footnotesize
\begin{tabular}{llccccc}
\toprule
\textbf{Supervision}&\textbf{Model} & \multicolumn{1}{c}{BERTScore} & \multicolumn{1}{c}{ROUGE} & \multicolumn{1}{c}{METEOR} & \multicolumn{1}{c}{SacreBLEU} & \multicolumn{1}{c}{BLEURT} \\ \midrule
3-shot prompting &GPT-4       & 92.49 & 34.31 & 41.58 & 6.04 & -31.02 \\ \midrule
\emph{SFT with LLM}& GPT2-XL & 92.45 & 33.52 & 37.13 & 6.35 & -39.81 \\ \midrule
\emph{SED with LLM}& GPT2-XL & 92.41 & 33.39 & 37.05 & 6.37 & -39.40  \\ \midrule
\emph{EaO with LLM}& GPT2-XL & 92.17 & 32.21 & 37.36 & 5.81 & -38.75 \\
\bottomrule
\end{tabular}
\caption{Automatic evaluation on $\alpha$NLG.}
\label{tab:anlg-auto}
\end{table*}

\begin{table*}[t]
\centering
\footnotesize
\begin{tabular}{llccccc}
\toprule
\textbf{Supervision}&\textbf{Model} & \multicolumn{1}{c}{BERTScore} & \multicolumn{1}{c}{ROUGE} & \multicolumn{1}{c}{METEOR} & \multicolumn{1}{c}{SacreBLEU} & \multicolumn{1}{c}{BLEURT} \\ \midrule
3-shot prompting &GPT-4 & 91.40 & 32.94 & 47.80 & 4.99 & -16.01 \\ \midrule
\emph{SFT with LLM} & GPT2-XL & 92.00 & 36.74 & 47.28 & 6.97 & -18.00 \\ \midrule
\emph{SED with LLM} & GPT2-XL & 91.89 & 36.22 & 46.58 & 5.80 & -18.57 \\ \midrule
\emph{EaO with LLM} & GPT2-XL & 91.89 & 36.07 & 47.41 & 5.88 & -16.68 \\

\bottomrule
\end{tabular}
\caption{Automatic evaluation on Sen-Making.}
\label{tab:sen-auto}
\end{table*}

\section{Templates}
\label{sec:template}
We include the following prompting templates:
\begin{itemize}
    \item Figure~\ref{fig:combine-question-answer-socialiqa}: The prompt to combine a question and its answer into a single sentence on \social with five demonstrations.
    \item Figure~\ref{fig:generate-improbable-answers}: The prompt to generate improbable answers on \social with six demonstrations.
    \item Figure~\ref{fig:estimate-outcome-likelihood}: The prompt to estimate the outcome likelihood given the context.
    \item Figure~\ref{fig:few-shot-prompting-explanation-socialiqa}: The prompt to generate explanations on \social with three demonstrations.
    \item Figure~\ref{fig:few-shot-prompting-explanation-rocstories}: The prompt to generate explanations on \story with three demonstrations.
    \item Figure~\ref{fig:improve-explanation}: The prompt to improve a crowd-written explanation.
\end{itemize}

\begin{figure*}
    \centering
    \includegraphics[width=\textwidth]{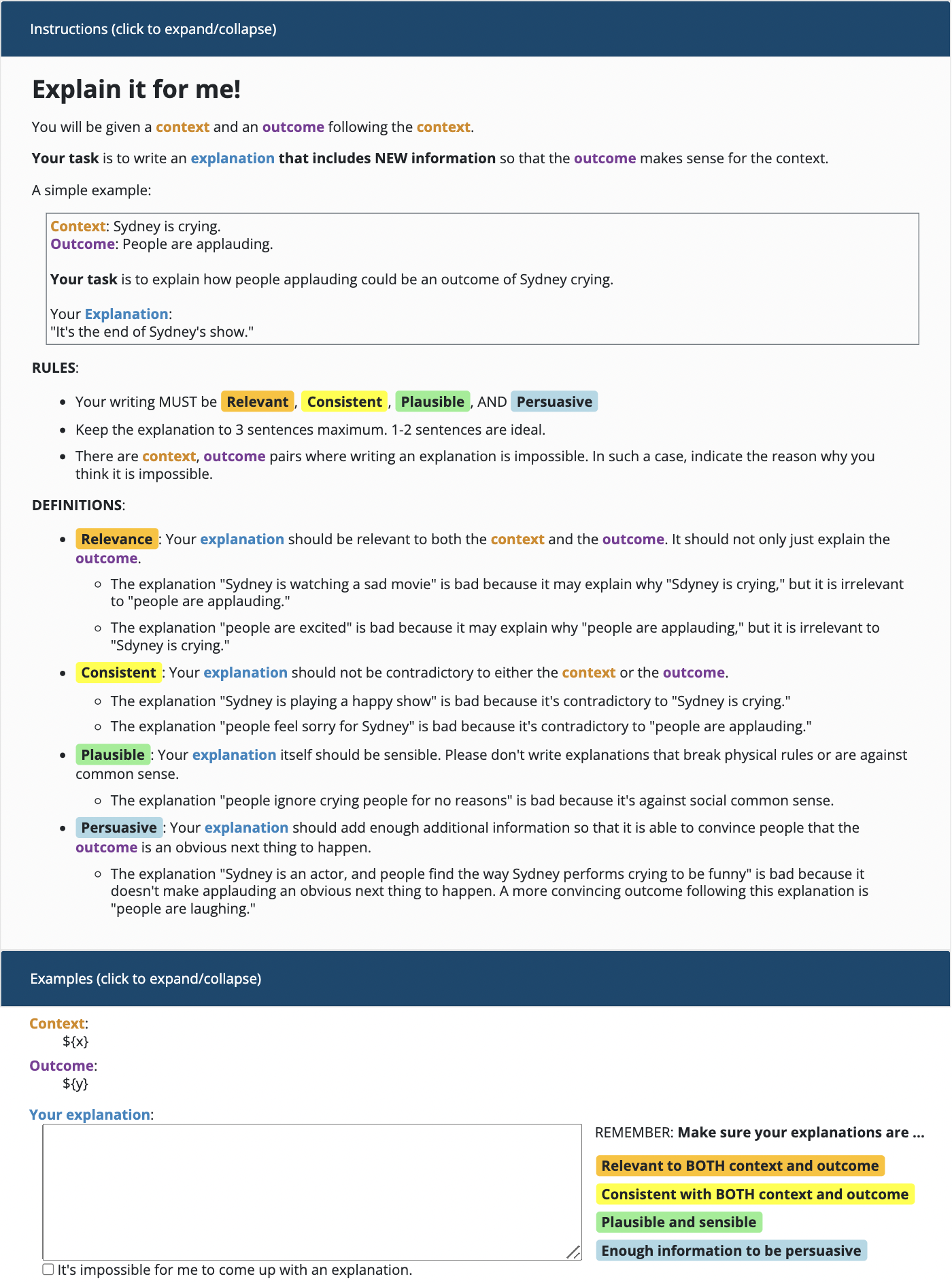}
    \caption{A screenshot of mturk template for collecting explanations.}
    \label{fig:mturk-explanation}
\end{figure*}

\begin{figure*}
    \centering
    \includegraphics[width=\textwidth]{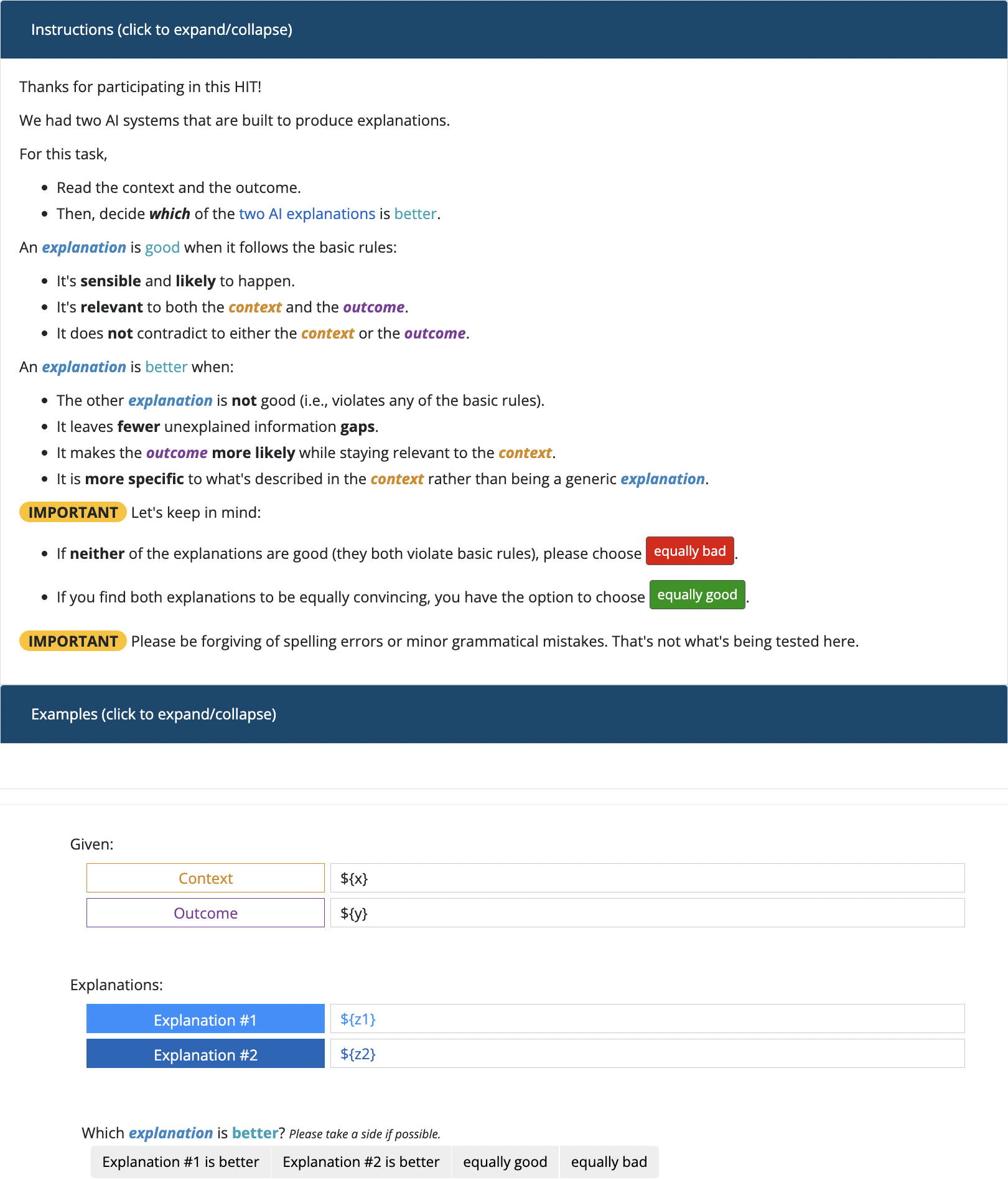}
    \caption{A screenshot of mturk template for doing pair-wise preference evaluation.}
    \label{fig:mturk-preference}
\end{figure*}

We also include the following MTurk templates:
\begin{itemize}
    \item Figure~\ref{fig:mturk-explanation}: The template to collect crowd-written explanations.
    \item Figure~\ref{fig:mturk-preference}: The template to collect human preferences.
\end{itemize}

\begin{figure}
    \centering
    \small
    \framebox{
    \parbox{0.45\textwidth}{
    Combine the following question and answer into a sentence: What will Others want to do next? quit their job and start their own business.\\
    Others will want to quit their job and start their own business. \smallskip \newline
    
    Combine the following question and answer into a sentence: How would you describe Remy? selfish\\
    Remy is selfish. \smallskip \newline
    
    Combine the following question and answer into a sentence: What will happen to Quinn? they will spontaneously combust\\
    Quinn will spontaneously combust. \smallskip \newline
    
    Combine the following question and answer into a sentence: How would you describe Bailey? do not want a healthy pet\\
    Bailey does not want a healthy pet. \smallskip \newline
    
    Combine the following question and answer into a sentence: How would you describe Carson? like Carson was mean\\
    Carson is mean. \smallskip \newline

    Combine the following question and answer into a sentence: \{question\} \{answer\}
    }
    }
    \caption{Prompting template for combining a question and its answer.}
    \label{fig:combine-question-answer-socialiqa}
\end{figure}
\begin{figure}
    \centering
    \small
    \framebox{
    \parbox{0.45\textwidth}{
    Context: Sydney walked past a homeless woman asking for change but did not have any money they could give to her. Sydney felt bad afterwards. \\
    Question: How would you describe Sydney? \\
    An unlikely answer: ridiculous \smallskip \newline

    Context: Jesse set Robin's suitcase on fire after their fight and messy breakup.\\
    Question: What will Jesse want to do next?\\
    An unlikely answer: decide not to reconcile\smallskip \newline
    
    Context: Bailey asked Sasha's grandma about church because they wanted to know more about it.\\
    Question: What will happen to Sasha?\\
    An unlikely answer: they get yelled by Sasha's grandma\smallskip \newline
    
    Context: Bailey told Alex to send the letter overnight since it was important.\\
    Question: How would Alex feel as a result?\\
    An unlikely answer: rushed\smallskip \newline
    
    Context: Lee made copies so that everyone at the table could follow along.\\
    Question: What will Lee want to do next?\\
    An unlikely answer: ask people stop reading the paper\smallskip \newline
    
    Context: Taylor gave Kai a lot to think about.\\
    Question: What will happen to Kai?\\
    An unlikely answer: not talk to Taylor\smallskip \newline
    
    Context: \{context\}\\
    Question: \{question\}\\
    An unlikely answer:
    }
    }
    \caption{Prompting template for generating improbable answers for SocialIQA examples.}
    \label{fig:generate-improbable-answers}
\end{figure}

\begin{figure}
    \centering
    \small
    \framebox{
    \parbox{0.45\textwidth}{
    A: \{context\}\\
    B: \{outcome\}\\
    On the scale from 1 to 5, how likely is B given A?
    }
    }
    \caption{Prompting template for estimating the likelihood of the outcome given the context.}
    \label{fig:estimate-outcome-likelihood}
\end{figure}

\begin{figure}
    \centering
    \small
    \framebox{
    \parbox{0.45\textwidth}{
    Context: Cameron decided to have a barbecue and gathered her friends together.\\
    Outcome: Others feel bored and uninterested.\\
    Explanation of the outcome: Other than eating the food, there weren't other activities planned.\smallskip \newline
    
    Context: Jan needed to give out jobs for an upcoming project at work.\\
    Outcome: Others will take a nap instead of working.\\
    Explanation of the outcome: The others don't get paid more for doing the jobs Jan gave out.\smallskip \newline
    
    Context: Remy was an expert fisherman and was on the water with Kai. Remy baited Kai's hook.\\
    Outcome: Remy will eat a sandwich.\\
    Explanation of the outcome: It's been too long before they feel the weight of a fish, and Remy is hungry.\smallskip \newline
    
    Context: \{context\}\\
    Outcome: \{outcome\}\\
    Explanation of the outcome:
    }
    }
    \caption{Prompting template for generating explanations for \social examples.}
    \label{fig:few-shot-prompting-explanation-socialiqa}
\end{figure}

\begin{figure}
    \centering
    \small
    \framebox{
    \parbox{0.45\textwidth}{
    Context: My friends all love to go to the club to dance. They think it's a lot of fun and always invite. I finally decided to tag along last Saturday. I danced terribly and broke a friend's toe.\\
    Outcome: My friends decided to keep inviting me out as I am so much fun.\\
    Explanation of the outcome: My friends thought the way I dance is really funny and they couldn't stop laughing.\smallskip \newline
    
    Context: On the fourth of July, Lilly baked a lemon blueberry cake. She brought it to her boyfriend's house and they had a bbq. After dinner they drove into the city to watch fireworks. When the show was over, they got donuts from a food truck.\\
    Outcome: Lilly had a terrible date.\\
    Explanation of the outcome: Lilly's boyfriend kept complaining that the donuts were way better than the lemon blueberry cake she made, and her boyfriend just threw the cake away.\smallskip \newline
    
    Context: Jennifer was bored one Saturday. She decided to alleviate her boredom with a hike. She drove to a national park to go hiking. Jennifer hiked for hours.\\
    Outcome: Jennifer thought hiking was stupid.\\
    Explanation of the outcome: She realized the Saturday was a holiday, and the hiking trails in the national park were too crowded that it took her longer than usual to finish.\smallskip \newline
    
    Context: \{context\}\\
    Outcome: \{outcome\}\\
    Explanation of the outcome:
    }
    }
    \caption{Prompting template for generating explanations for \story examples.}
    \label{fig:few-shot-prompting-explanation-rocstories}
\end{figure}

\begin{figure}[t]
    \centering
    \small
    \framebox{
    \parbox{0.45\textwidth}{
    Can you improve this explanation so that it becomes more specific to the context and makes the outcome more likely to happen?\\ \smallskip \newline
    Context: \{context\}\\
    Outcome: \{outcome\}\\
    Explanation for the outcome:\{explanation\}
    }
    }
    \caption{Prompting template for improving an explanation.}
    \label{fig:improve-explanation}
\end{figure}


\section{Crowdsourcing Details}
\label{app:crowdsourcing}

Tasks which are allocated to a worker but not completed are later distributed to the entire group of workers.
We allow workers at least a week to complete each of their allocated tasks, which allows them sufficient time to complete the task and work at their own pace.


\subsection{Qualification.}
We use a qualification task to recruit and train workers to produce quality explanations of uncommon outcomes. 
In the qualification task, each worker is asked to write an explanation for five pre-chosen contexts paired with uncommon outcomes, including one pair chosen as an attention check.
Three paper authors manually grade the explanations to check if they make the outcomes more likely, naturally follow the contexts, and leave little information gaps in-between. 
We qualify workers who provide at least three high-quality explanations, resulting in qualifying 204 out of 520 workers.

\subsection{Quality Control for Crowd-written Explanations.}
To ensure the quality of crowd-written explanations, we maintain active communication with workers, and detect and filter low-quality explanations.
We engage with workers through an online group chat and periodically provide personalized feedback to individual workers.
We detect low-quality explanations through multiple manual and automatic filters, e.g., checking for contradictions between the worker-written explanation and the context and outcome. 
We dequalify 22 workers who contribute more than two low-quality out of five randomly sampled explanations, and remove all of their explanations from the dataset.

Additionally, we have following automatic ways to verify workers' explanations:
\begin{itemize}
    \item We use GPT3 to check contradiction between a context and its corresponding explanations.
    \item We use GPT2 to check relevance to the context via $p(y|x, z)-p(y|z) > \epsilon$.
    \item In each launch, we sample one explanation from each worker, and we send individual feedback to the workers who violate our rules and filter out the workers who contributed bad explanations to us 
    \item We check how many examples are marked impossible to explain by each worker, and remove workers who use such marks too often. 
\end{itemize}

\section{Experiment Model Details}
\label{app:experiment_details}
We implement both the baseline and the proposed approaches with Hugging Face Transformers~\cite{wolf-etal-2020-transformers}. We train all models with a learning rate of 0.00001 and a batch size of 8. We perform grid search with $\lambda \in \{1, 0.1, 0.01\}$ and $\beta \in \{0.1, 0.01, 0.001\}$, and we choose the best performing checkpoint on the development set.
In DAgger, we set epochs $I$ to be five and block size $k$ to be 2 tokens.

\section{Static expert demonstrations pseudo-code}
\label{sec:sec_with_static_pseduocode}

The pseduocode for the static expert demonstrations algorithm introduced in \S\ref{sec:sec_with_static_expert_demos} is given in Algorithm~\ref{alg:seqnll}.

\begin{algorithm}[t]
\footnotesize
\caption{Online learning with static expert demonstrations.}
\label{alg:seqnll}
\begin{algorithmic}[1]
\State \textbf{Inputs:} Initial learner policy parameters $\theta_0$, dataset $\mathcal{D}=\{(x, y, z)\}^N$, number of training epochs $I$.
\State $\tilde{D} \gets \emptyset$ 
\For{$i=0,\dotsc, I - 1$}
    \For{$(x, y, z) \in \mathcal{D}$}
        \State $\tilde{z} \ \sim \pi(. \ | \ x, y)$
        \State $\tilde{\mathcal{D}} \gets \tilde{\mathcal{D}} \cup \{(x, y, z, \tilde{z})\}$
    \EndFor
    \State $\theta_{i + 1} \gets \theta_i$ further optimized on $\tilde{D}$ using the objective in Equation~\ref{eq:seqnll}.
\EndFor
\State \textbf{Returns:} Learned policy parameters $\theta_I$.
\end{algorithmic}
\end{algorithm}

\end{document}